\documentclass{article}
\PassOptionsToPackage{sort&compress,numbers}{natbib}
\usepackage[final]{neurips_2020}
\usepackage[utf8]{inputenc}
\usepackage[T1]{fontenc}
\usepackage{graphics,graphicx,caption,float,subcaption,booktabs,xcolor,multirow,array,color,ifthen,tabu,colortbl,dblfloatfix,url,xparse,mathtools,patchcmd,algorithm,algorithmic,amssymb,xspace,nicefrac,microtype,amsmath,amsfonts,bm,ragged2e,tikz,stackengine,etoolbox,xpatch,enumerate,xstring,setspace,tabularx,makecell,changepage,cuted,titlesec,enumitem,wrapfig,tcolorbox}
\usepackage[pagebackref=true,breaklinks=true,colorlinks=true,bookmarks=false,citecolor=blue]{hyperref}
\usepackage[capitalise,noabbrev,nameinlink]{cleveref}
\usepackage[english]{babel}
\definecolor{forestgreen}{rgb}{0.13, 0.55, 0.13}
\definecolor{coralpink}{rgb}{0.97, 0.51, 0.47}
\definecolor{coralred}{rgb}{1.0, 0.25, 0.25}
\definecolor{lavenderpink}{rgb}{0.98, 0.68, 0.82}

\newcommand{\our}{NDP\xspace}
\newcommand{\ours}{NDPs\xspace}

\title{Neural Dynamic Policies\\for End-to-End Sensorimotor Learning}
\author{Shikhar Bahl\\
CMU\\
\And
Mustafa Mukadam \\
FAIR\\
\And
Abhinav Gupta \\
CMU\\
\And
Deepak Pathak \\
CMU\\
}

\begin{document}
\maketitle

\begin{abstract}
\begin{hyphenrules}{nohyphenation}
The current dominant paradigm in sensorimotor control, whether imitation or reinforcement learning, is to train policies directly in raw action spaces such as torque, joint angle, or end-effector position. This forces the agent to make decisions individually at each timestep in training, and hence, limits the scalability to continuous, high-dimensional, and long-horizon tasks. In contrast, research in classical robotics has, for a long time, exploited dynamical systems as a policy representation to learn robot behaviors via demonstrations. These techniques, however, lack the flexibility and generalizability provided by deep learning or reinforcement learning and have remained under-explored in such settings. In this work, we begin to close this gap and embed the structure of a dynamical system into deep neural network-based policies by reparameterizing action spaces via second-order differential equations. We propose Neural Dynamic Policies (NDPs) that make predictions in trajectory distribution space as opposed to prior policy learning methods where actions represent the raw control space. The embedded structure allows end-to-end policy learning for both reinforcement and imitation learning setups. We show that NDPs outperform the prior state-of-the-art in terms of either efficiency or performance across several robotic control tasks for both imitation and reinforcement learning setups. Project video and code are available at: \url{https://shikharbahl.github.io/neural-dynamic-policies/}.
\end{hyphenrules}
\end{abstract}

\section{Introduction}
\label{intro}
\begin{wrapfigure}{r}{0.46\textwidth}
  \vspace{-6mm}
  \begin{center}
    \includegraphics[width=\linewidth]{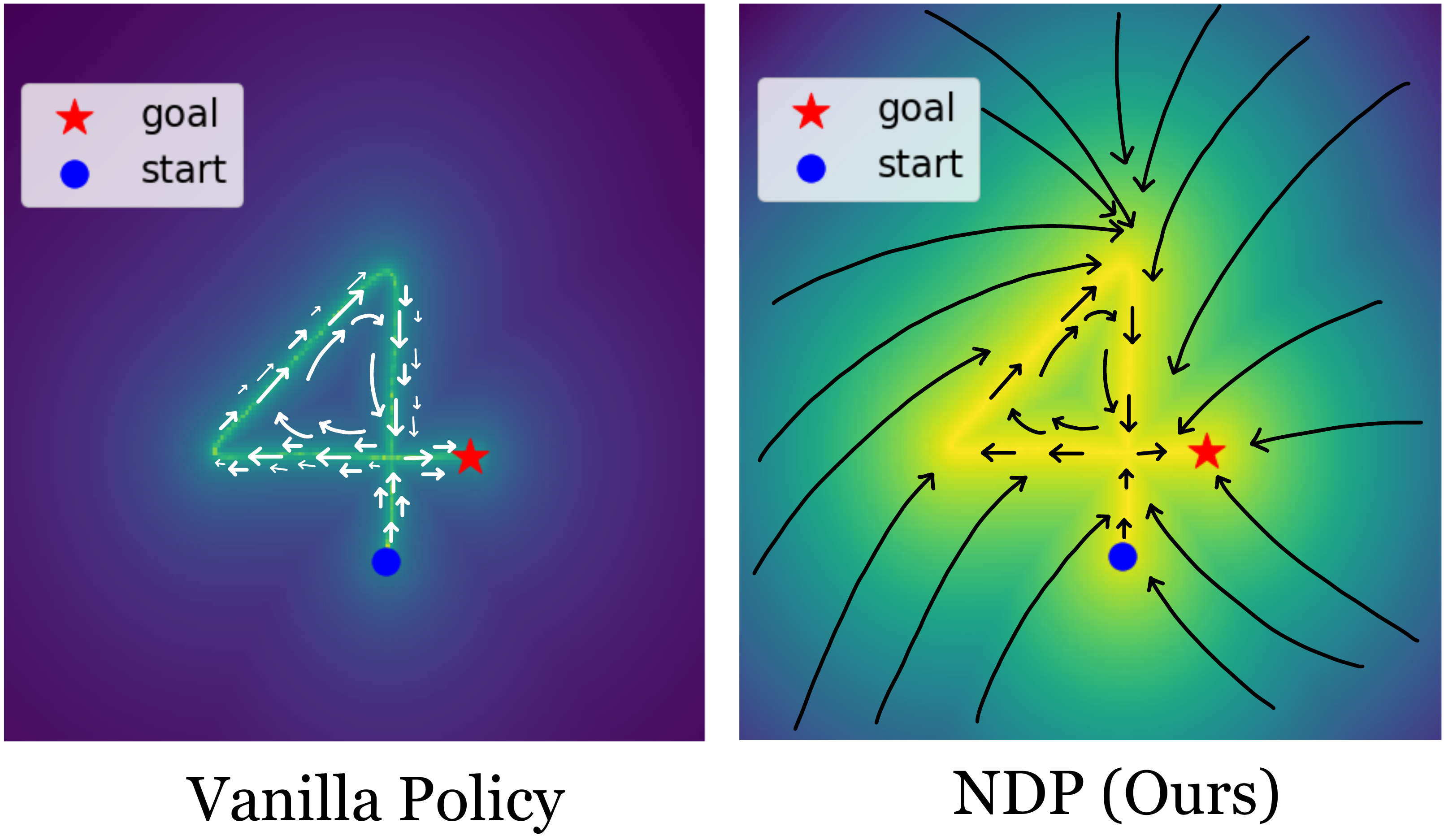}
  \end{center}
  \vspace{-0.1in}
  \caption{\small Vector field induced by NDPs. The goal is to draw the planar digit 4 from the start position. The dynamical structure in NDP induces a smooth vector field in trajectory space. In contrast, a vanilla policy has to reason individually in different parts.}
  \label{fig:teaser}
\end{wrapfigure}
Consider an embodied agent tasked with throwing a ball into a bin. Not only does the agent need to decide where and when to release the ball, but also needs to reason about the whole trajectory that it should take such that the ball is imparted with the correct momentum to reach the bin. This form of reasoning is necessary to perform many such everyday tasks. Common methods in deep learning for robotics tackle this problem either via imitation or reinforcement learning. However, in most cases, the agent's policy is trained in raw action spaces like torques, joint angles, or end-effector positions, which forces the agent to make decisions at each time step of the trajectory instead of making decisions in the trajectory space itself (see Figure~\ref{fig:teaser}). But then how do we reason about trajectories as actions?

A good trajectory parameterization is one that is able to capture a large set of an agent's behaviors or motions while being physically plausible. In fact, a similar question is also faced by scientists while modeling physical phenomena in nature. Several systems in science, ranging from motion of planets to pendulums, are described by differential equations of the form $\ddot{{y}} = m^{-1}f(y,\dot{y})$, where $y$ is the generalized coordinate, $\dot{y}$ and $\ddot{y}$ are time derivatives, $m$ is mass, and $f$ is force. Can a similar parameterization be used to describe the behavior of a robotic agent? Indeed, classical robotics has leveraged this connection to represent task specific robot behaviors for many years. In particular, dynamic movement primitives (DMP)~\cite{schaal2006dynamic,isprt2012dmp,ijspeert2002movement,ijspeert2003learning} have been one of the more prominent approaches in this area. Despite their successes, DMPs have not been explored much beyond behavior cloning paradigms. This is partly because these methods tend to be sensitive to parameter tuning and aren't as flexible or generalizable as current end-to-end deep network based approaches.

In this work, we propose to bridge this gap by embedding the structure of dynamical systems\footnote{Dynamical systems here should not be confused with dynamics model of the agent. We incorporate dynamical differential equations to represent robot's behavioral trajectory and not physical transition dynamics.} into deep neural network-based policies such that the agent can directly learn in the space of physically plausible trajectory distributions (see Figure~\ref{fig:teaser}(b)). \textbf{Our key insight is to reparameterize the action space in a deep policy network with nonlinear differential equations corresponding to a dynamical system and train it end-to-end over time in either reinforcement learning or imitation learning setups.}
However, this is quite challenging to accomplish, since naively predicting a full arbitrary dynamical system directly from the input, trades one hard problem for another. Instead, we want to prescribe some structure such that the dynamical system itself manifests as a layer in the deep policy that is both, amenable to take arbitrary outputs of previous layers as inputs, and is also fully differentiable to allow for gradients to backpropagate.

We address these challenges through our approach, Neural Dynamic Policies (\ours). Specifically, \ours allow embedding desired dynamical structure as a layer in deep networks. The parameters of the dynamical system are then predicted as outputs of the preceding layers in the architecture conditioned on the input. \textbf{The `deep' part of the policy then only needs to reason in the lower-dimensional space of building a dynamical system that then lets the overall policy easily reason in the space of trajectories.} In this paper, we employ the aforementioned DMPs as the structure for the dynamical system and show its differentiability, although they only serve as a design choice and can possibly be swapped for a different differentiable dynamical structure, such as RMPs~\cite{ratliff2018riemannian}.

We evaluate \ours in imitation as well as reinforcement learning setups. \ours can utilize high-dimensional inputs via demonstrations and learn from weak supervisory signals as well as rewards. In both settings, \ours exhibit better or comparable performance to state-of-the-art approaches.

\begin{figure}[t]
  \centering
  \includegraphics[width=.95\linewidth]{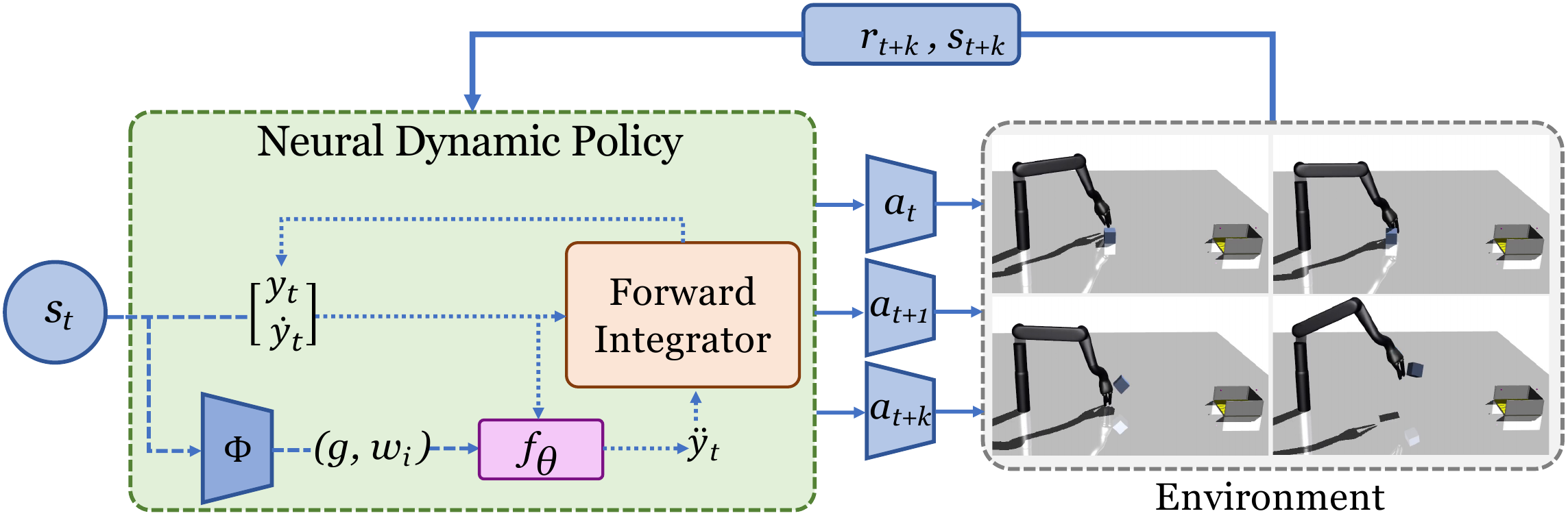}
  \caption{\small Given an observation from the environment, $s_t$, a Neural Dynamic Policy generates parameters $w$ (weights of basis functions) and $g$ (goal for the robot) for a forcing function $f_\theta$. An open loop controller then uses this function to output a set of actions for the robot to execute in the environment, collecting future states and rewards to train the policy.}
  \vspace{-0.2in}
  \label{fig:method}
\end{figure}

\section{Modeling Trajectories with Dynamical Systems}
\label{sec:dmp}
Consider a robotic arm exhibiting a certain behavior to accomplish some task. Given a choice of coordinate system, such as either joint-angles or end-effector position, let the state of the robot be $y$, velocity $\dot{y}$ and acceleration $\ddot{y}$. In mechanics, Euler-Lagrange equations are used to derive the equations of motion as a general second order dynamical system that perfectly captures this behavior~\cite[Chapter~6]{spong2020robot}. It is common in classical robotics to represent movement behaviors with such a dynamical system. Specifically, we follow the second order differential equation structure imposed by Dynamic Movement Primitives~\cite{isprt2012dmp,schaal2006dynamic,pastor2009motorskills}. Given a desired goal state $g$, the behavior is represented as:
\begin{equation}
    \ddot{y} = \alpha(\beta(g - y) - \dot{y}) + f(x),
    \label{eq:dmp}
\end{equation}
where $\alpha, \beta$ are global parameters that allow critical damping of the system and smooth convergence to the goal state. $f$ is a non-linear forcing function which captures the shape of trajectory and operates over $x$ which serves to replace time dependency across trajectories, giving us the ability to model time invariant tasks, e.g., rhythmic motions. $x$ evolves through the first-order linear system:
\begin{equation}
    \dot{x} = -a_x x
    \label{eq:canonical}
\end{equation}
 The specifics of $f$ are usually design choices. We use a sum of weighted Gaussian radial basis functions \cite{isprt2012dmp} shown below:
 \begin{equation}
    f(x, g) = \frac{\sum \psi_i w_i}{\sum \psi_i}x(g - y_0), \quad
    \psi_i = e^{(-h_i(x-c_i)^2)}
    \label{eq:forcing_func}
\end{equation}
where $i$ indexes over $n$ which is the number of basis functions. Coefficients $c_i = e^{\frac{-i\alpha_x}{n}}$ are the horizontal shifts of each basis function, and $h_i = \frac{n}{c_i}$ are the width of each of each basis function. The weights on each of the basis functions $w_i$ parameterize the forcing function $f$. This set of nonlinear differential equations induces a smooth trajectory distribution that acts as an attractor towards a desired goal (see Figure~\ref{fig:teaser}, right). We now discuss how to combine this dynamical structure with deep neural network based policies in an end-to-end differentiable manner.

\section{Neural Dynamic Policies (\ours)}
We condense actions into a space of trajectories, parameterized by a dynamical system, while keeping all the advantages of a deep learning based setup. We present a type of policy network, called Neural Dynamic Policies (\ours) that given an input, image or state, can produce parameters for an embedded dynamical structure, which reasons in trajectory space but outputs raw actions to be executed. Let the unstructured input to robot be $s$, (an image or any other sensory input), and the action executed by the robot be $a$. We describe how we can incorporate a dynamical system as a differentiable layer in the policy network, and how \ours can be utilized to learn complex agent behaviors in both imitation and reinforcement learning settings.

\subsection{Neural Network Layer Parameterized by a Dynamical System}
Throughout this paper, we employ the dynamical system described by the second order DMP equation~\eqref{eq:dmp}. There are two key parameters that define what behavior will be described by the dynamical system presented in Section~\ref{sec:dmp}: basis function weights $w=\{w_1, \dots, w_i, \dots, w_n\}$ and goal $g$. NDPs employ a neural network $\Phi$ which takes an unstructured input $s$\footnote{robot's state $y$ is not to be confused with environment observation $s$ which contains world as well as robot state (and often velocity). $s$ could be given by either an image or true state of the environment.} and predicts the parameters $w, g$ of the dynamical system. These predicted $w,g$ are then used to solve the second order differential equation~\eqref{eq:dmp} to obtain system states $\{y,\dot{y},\ddot{y}\}$. Depending on the difference between the choice of robot's coordinate system for $y$ and desired action $a$, we may need an inverse controller $\Omega(.)$ to convert $y$ to $a$, i.e., $a=\Omega(y,\dot{y},\ddot{y})$. For instance, if $y$ is in joint angle space and $a$ is a torque, then $\Omega(.)$ is the robot's inverse dynamics controller, and if $y$ and $a$ both are in joint angle space then $\Omega(.)$ is the identity function.

As summarized in Figure~\ref{fig:method}, neural dynamic policies are defined as $\pi(a|s; \theta) \triangleq \Omega\big(\texttt{DE}\big(\Phi(s; \theta)\big)\big)$ where $\texttt{DE}(w,g) \rightarrow \{y,\dot{y},\ddot{y}\}$ denotes solution of the differential equation~\eqref{eq:dmp}. The forward pass of $\pi(a|s)$ involves solving the dynamical system and backpropagation requires it to be differentiable. We now show how we differentiate through the dynamical system to train the parameters $\theta$ of \ours.

\subsection{Training NDPs by Differentiating through the Dynamical System}
\label{sec:dmp-train}
To train NDPs, estimated policy gradients must flow from $a$, through the parameters of the dynamical system $w$ and $g$, to the network $\Phi(s; \theta)$. At any time $t$, given the previous state of robot $y_{t-1}$ and velocity $\dot{y}_{t-1}$ the output of the DMP in Equation~\eqref{eq:dmp} is given by the acceleration
\begin{equation}
\ddot{y}_t = \alpha(\beta(g - y_{t - 1}) - \dot{y}_{t - 1} + f(x_t, g)
\label{eq:accel_dmp_diff}
\end{equation}
Through Euler integration, we can find the next velocity and position after a small time interval $dt$
\begin{equation}
\dot{y}_t = \dot{y}_{t - 1} + \ddot{y}_{t - 1}dt, \quad y_t = y_{t - 1} + \dot{y}_{t - 1}dt
\label{eq:dmp_diff}
\end{equation}
In practice, this integration is implemented in $m$ discrete steps. To perform a forward pass, we unroll the integrator for $m$ iterations starting from initial $\dot{y}_0$, $\ddot{y}_0$.
We can either apply all the $m$ intermediate robot states $y$ as actions on the robot using inverse controller $\Omega(.)$, or equally sub-sample them into $k\in\{1,m\}$ actions in between, where $k$ is the \our rollout length. This frequency of sampling could allow robot operation at a much higher frequency (.5-5KHz) than the environment (usually 100Hz). The sampling frequency need not be same at training and inference as discussed further in Section~\ref{sec:dmp-infer}.

Now we can compute gradients of the trajectory from the DMP with respect to $w$ and $g$ using Equations~\eqref{eq:forcing_func}-\eqref{eq:dmp_diff} as follows:
\begin{equation}
\frac{\partial{f(x_t, g)}}{\partial{w_i}} = \frac{\psi_i}{\sum_j \psi_j}(g - y_0)x_t, \quad \frac{\partial{f(x_t, g)}}{\partial{g}} = \frac{\psi_j w_j}{\sum_j \psi_j}x_t
\label{eq:dmp_diff_final}
\end{equation}
Using this, a recursive relationship follows between, (similarly to the one derived by ~\citet{pahic2018deepenc}) $\frac{\partial{y_t}}{\partial{w_i}}$, $\frac{\partial{y_t}}{\partial{g}}$ and the preceding derivatives of $w_i$, $g$ with respect to $y_{t - 1}$, $y_{t - 2}$, $\dot{y}_{t-1}$ and $\dot{y}_{t-2}$. Complete derivation of equation~\eqref{eq:dmp_diff_final} is given in appendix.

We now discuss how \ours can be leveraged to train policies for imitation learning and reinforcement learning setups.

\subsection{Training NDPs for Imitation (Supervised) Learning}
\label{sec:dmp-sl}
Training NDPs in imitation learning setup is rather straightforward. Given a sequence of input $\{s,s',\dots\}$, \our's $\pi(s; \theta)$ outputs a sequence of actions {${a, a' \dots}$}. In our experiments, $s$ is a high dimensional image input. Let the demonstrated action sequence be $\tau_\text{target}$, we just take a loss between the predicted sequence as follows:

\begin{equation}
    \mathcal{L}_\text{imitation} = \sum_{s} ||\pi(s) -  \tau_\text{target}(s)||^2
\end{equation}
The gradients of this loss are backpropagated as described in Section 3.2 to train the parameters $\theta$.

\subsection{Training NDPs for Reinforcement Learning}
\label{sec:dmp-rl}
\begin{wrapfigure}{r}{0.4\textwidth}
\vspace{-14mm}
\begin{minipage}{0.475\textwidth}
\begin{algorithm}[H]
   	\footnotesize
   	\caption{Training NDPs for RL}
   	\begin{algorithmic}
    \REQUIRE Policy $\pi$, $k$ NDP rollout length, $\Omega$ low-level inverse controller
    \FOR{$1, 2, ...$ episodes}
        \FOR {$t = 0, k, \dots, $ until end of episode}
            \STATE $w$, $g$ = $\Phi (s_t)$ \
            \STATE Robot $y_t$, $\dot{y_t}$ from $s_t$ (pos, vel)
            \FOR {$m = 1, ..., M$ (integration steps)}
                \STATE Estimate $\dot{x}_m$ via \eqref{eq:canonical} and update $x_m$
                \STATE Estimate $\ddot{y}_{t + m}$, $\dot{y}_{t + m}$, $y_{t + m}$ via \eqref{eq:accel_dmp_diff},  \eqref{eq:dmp_diff}
                \STATE $a = \Omega(y_{t + m}, y_{t + m - 1})$
                \STATE Apply action $a$ to get $s'$
                \STATE Store transition ($s, a, s', r$)
            \ENDFOR
        \STATE Compute Policy gradient $\nabla_\theta J$
        \STATE $\theta  \leftarrow \theta + \eta \nabla_\theta J $
        \ENDFOR
    \ENDFOR
   	\end{algorithmic}
   	\label{summary}
\end{algorithm}
\end{minipage}
\vspace{-5mm}
\end{wrapfigure}
We now show how an \our can be used as a policy, $\pi$ in the RL setting. As discussed in Section~\ref{sec:dmp-train}, \our samples k actions for the agent to execute in the environment given input observation $s$. One could use any underlying RL algorithm to optimize the expected future returns. In this paper, we use Proximal Policy Optimization (PPO) \cite{ppo} and treat $a$ independently when computing the policy gradient for each step of the \our rollout and backprop via a reinforce objective.

There are two choices for value function critic $V^\pi(s)$: either predict a single common value function for all the actions in the $k$-step rollout or predict different critic values for each step in the \our rollout sequence. We found that the latter works better in practice. We call this a \textit{multi-action critic architecture} and predict $k$ different estimates of value using $k$-heads on top of the critic network. Later, in the experiments we perform ablations over the choice of $k$. To further create a strong baseline comparison, as we discuss in Section~\ref{sec:baseline}, we also design and compare against a variant of PPO that predicts multiple actions using our multi-action critic architecture.

Algorithm~\ref{summary} provides a summary of our method for training \ours with policy gradients. We only show results of using \ours with on-policy RL (PPO), however, \ours can similarly be adapted to off-policy methods.

\subsection{Inference in NDPs}
\label{sec:dmp-infer}
In the case of inference, our method uses the \our policy $\pi$ once every $k$ environment steps, hence requires $k$-times fewer forward passes as actions applied to the robot. While reducing the inference time in simulated tasks may not show much difference, in real world settings, where large perception systems are usually involved, reducing inference time can help decrease overall time costs. Additionally, deployed real world systems may not have the same computational power as many systems used to train state-of-the-art RL methods on simulators, so inference costs end up accumulating, thus a method that does inference efficiently can be beneficial. Furthermore, as discussed in Section~\ref{sec:dmp-train}, the rollout length of \our can be more densely sampled at test-time than at training allowing the robot to produce smooth and dynamically stable motions. Compared to about 100Hz frequency of the simulation, our method can make decisions an order of magnitude faster (at about 0.5-5KHz) at inference.

\begin{figure}[t!]
\centering
\begin{subfigure}[b]{0.155\linewidth}
    \frame{\includegraphics[width=\linewidth]{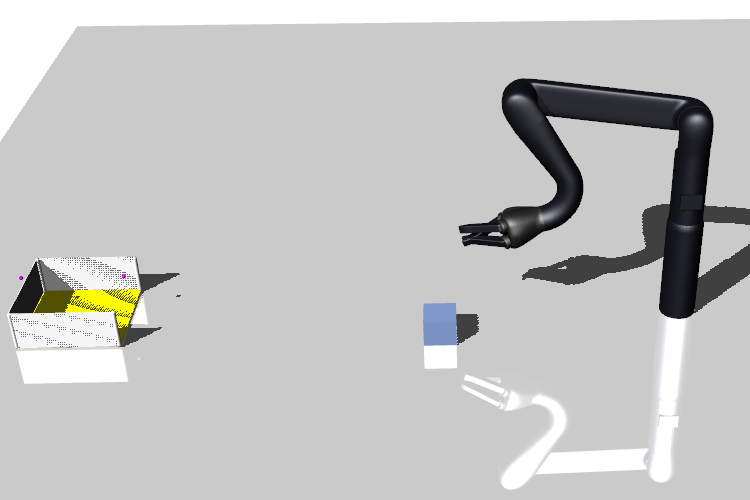}}
    \vspace{-0.18in}
    \caption{\small Throwing}
    \label{fig:throw-env}
\end{subfigure}
\begin{subfigure}[b]{0.155\linewidth}
    \frame{\includegraphics[width=\linewidth]{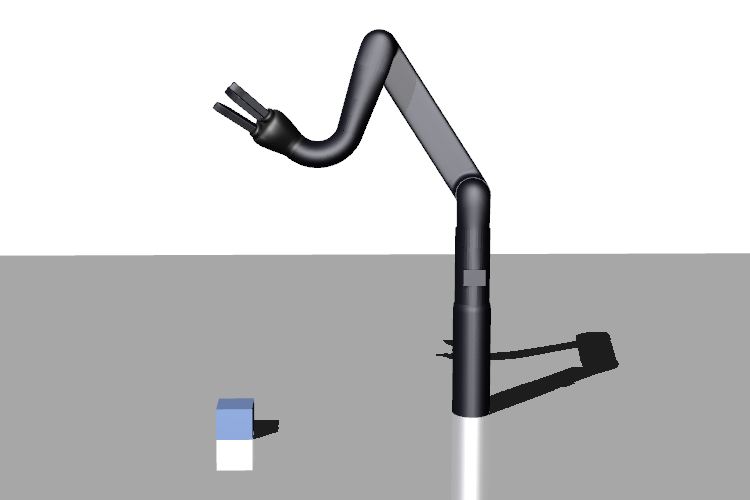}}
    \vspace{-0.18in}
    \caption{\small Picking}
    \label{fig:pick-env}
\end{subfigure}
\begin{subfigure}[b]{0.155\linewidth}
    \frame{\includegraphics[width=\linewidth]{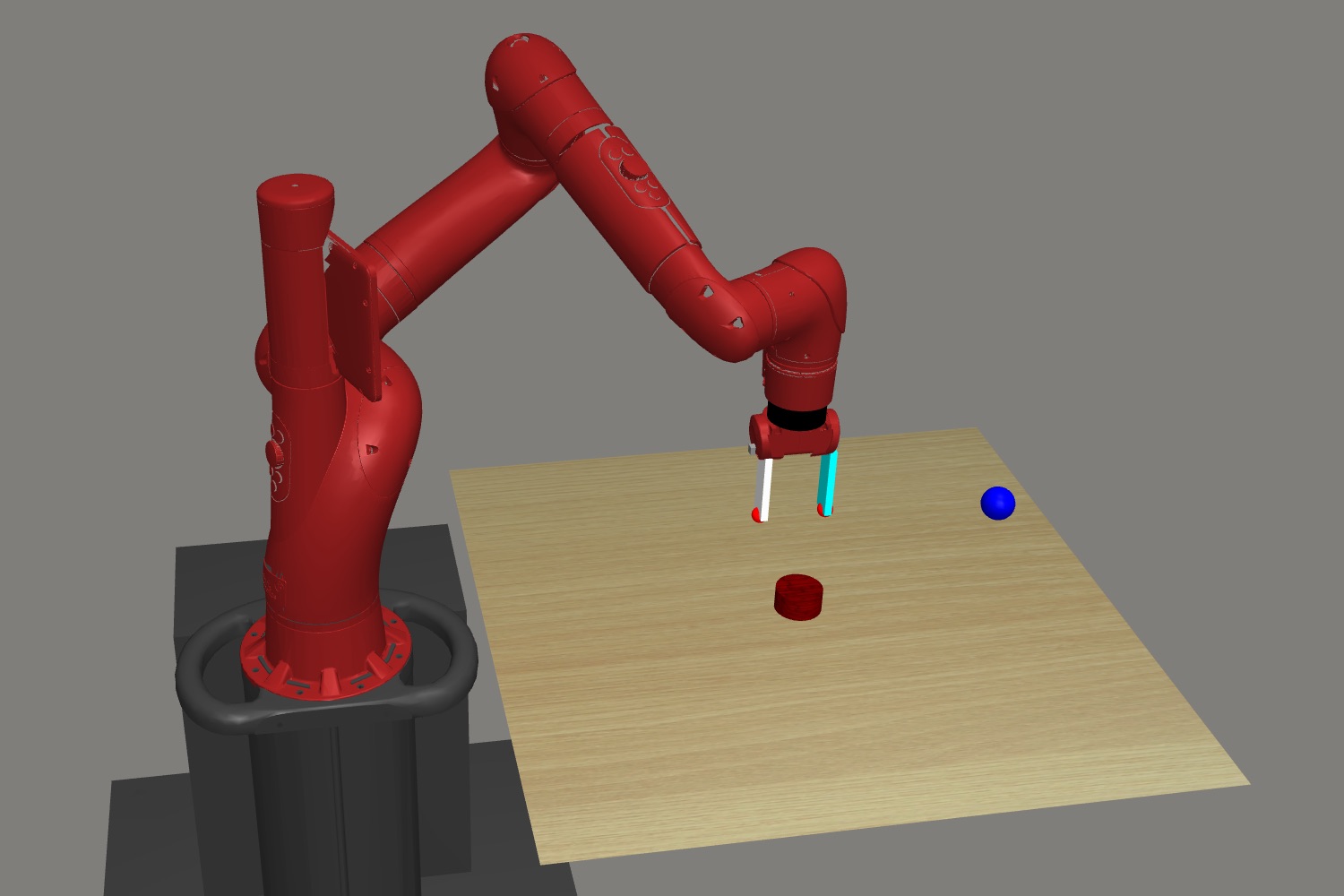}}
    \vspace{-0.18in}
    \caption{\small Pushing}
    \label{fig:push-env}
\end{subfigure}
\begin{subfigure}[b]{0.155\linewidth}
    \frame{\includegraphics[width=\linewidth]{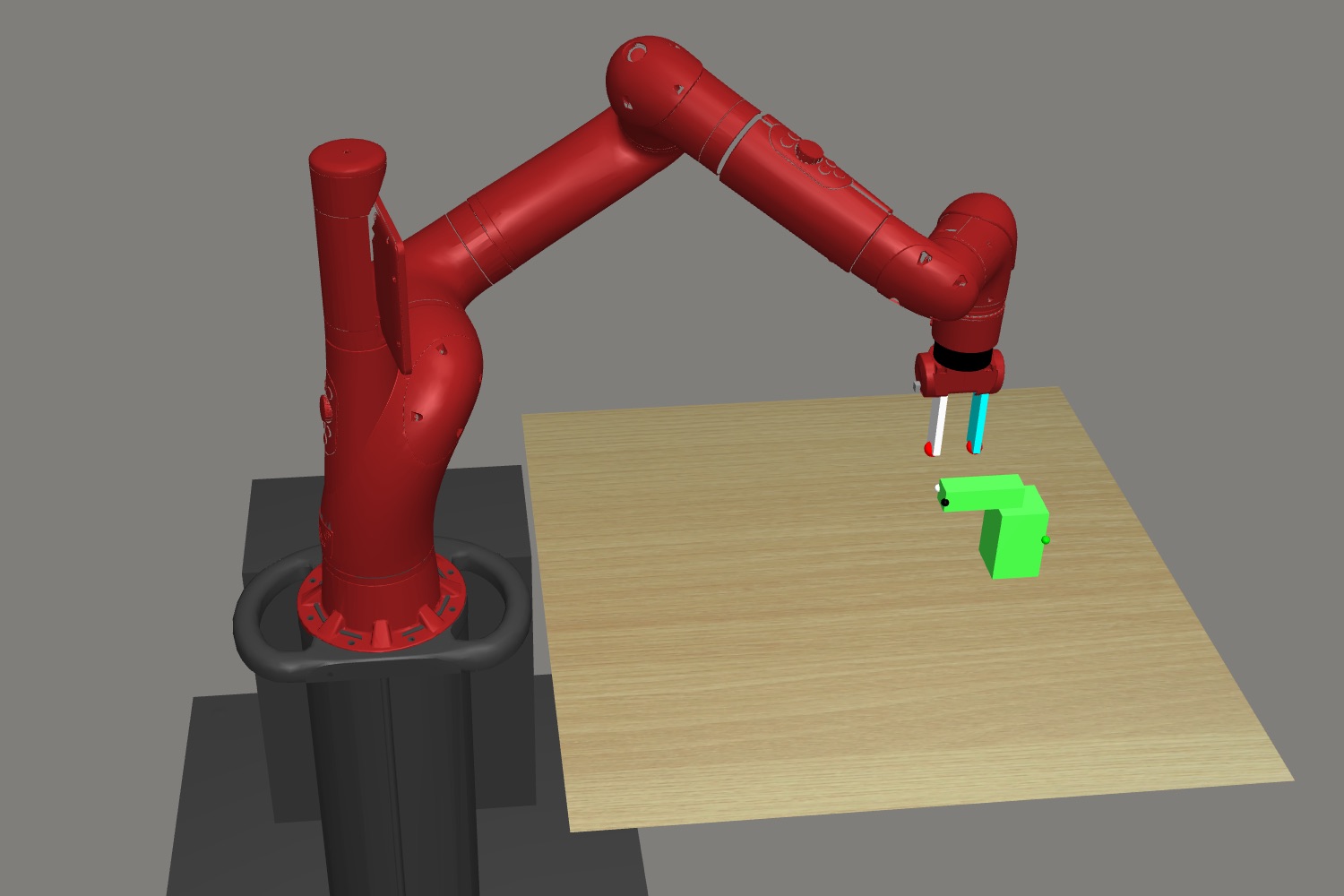}}
    \vspace{-0.18in}
    \caption{\small Faucet Open}
    \label{fig:faucet-env}
\end{subfigure}
\begin{subfigure}[b]{0.155\linewidth}
    \frame{\includegraphics[width=\linewidth]{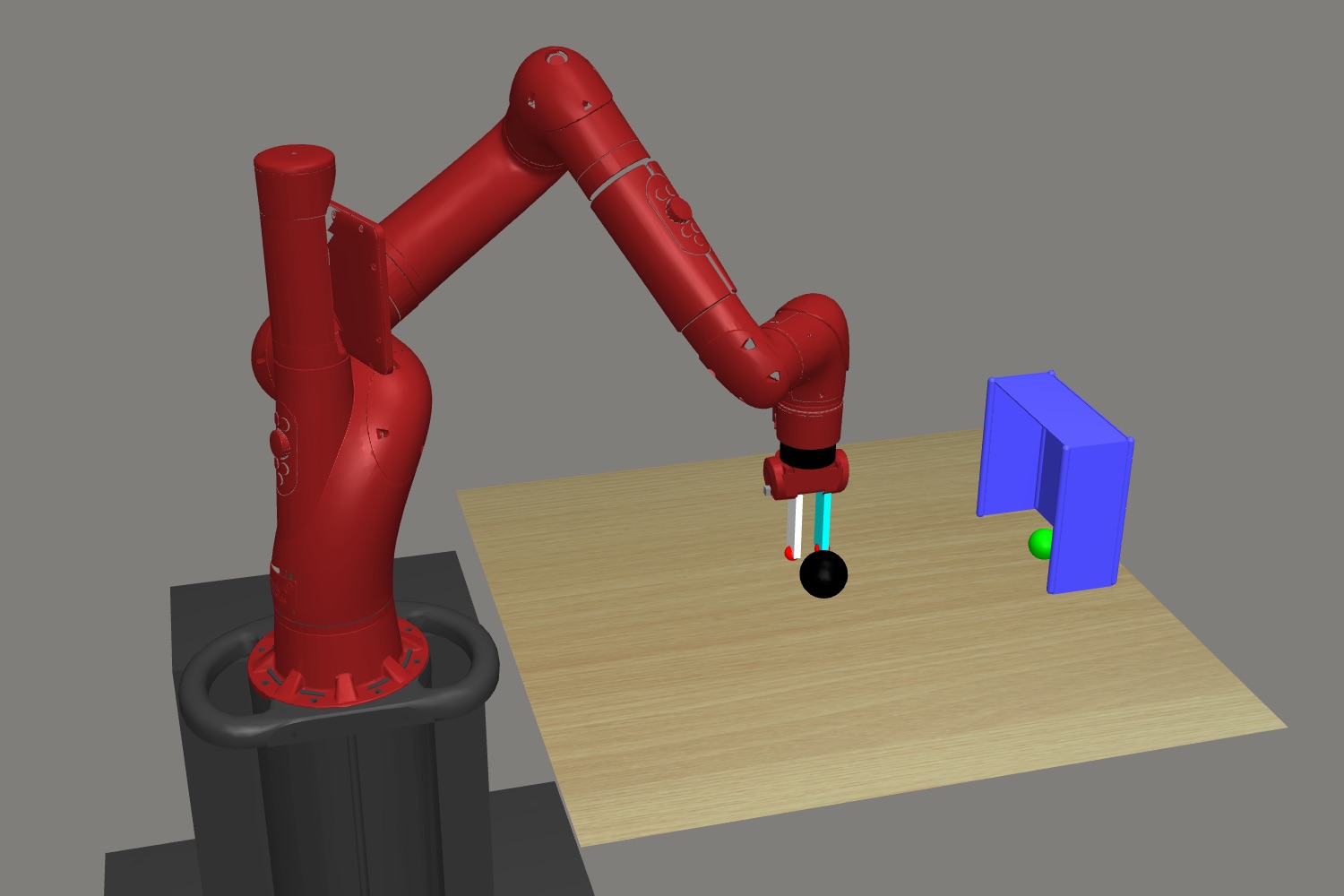}}
    \vspace{-0.18in}
    \caption{\small Soccer}
    \label{fig:soccer-env}
\end{subfigure}
\begin{subfigure}[b]{0.184\linewidth}
    \frame{\includegraphics[width=\linewidth]{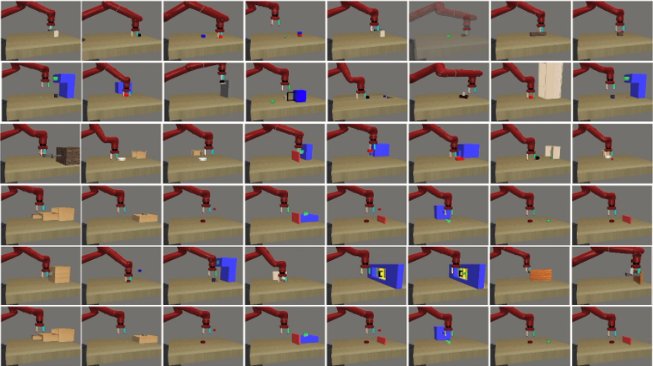}}
    \vspace{-0.18in}
    \caption{\small 50 Tasks}
    \label{fig:door-env}
\end{subfigure}
\vspace{-0.06in}
\caption{\small Environment snapshot for different tasks considered in experiments. (a,b) Throwing and Picking tasks are adapted from \cite{ghosh2017divide} on the Kinova Jaco arm. (c-f) Remaining tasks are adapted from \citet{yu2019meta}.}
\vspace{-0.01in}
\label{fig:rl-envs}
\end{figure}

\begin{figure}[t!]
\centering
\begin{subfigure}[b]{0.115\linewidth}
    \frame{\includegraphics[width=\linewidth]{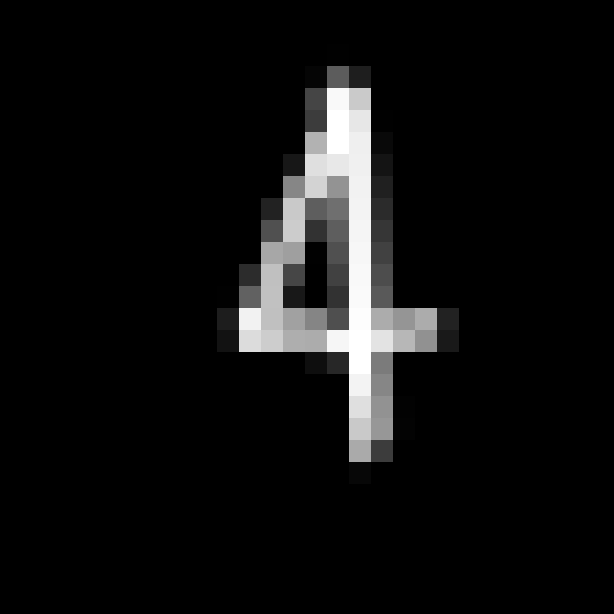}}
    \vspace{-0.2in}
    \caption*{\small Input}
\end{subfigure}
\begin{subfigure}[b]{0.115\linewidth}
    \frame{\includegraphics[width=\linewidth]{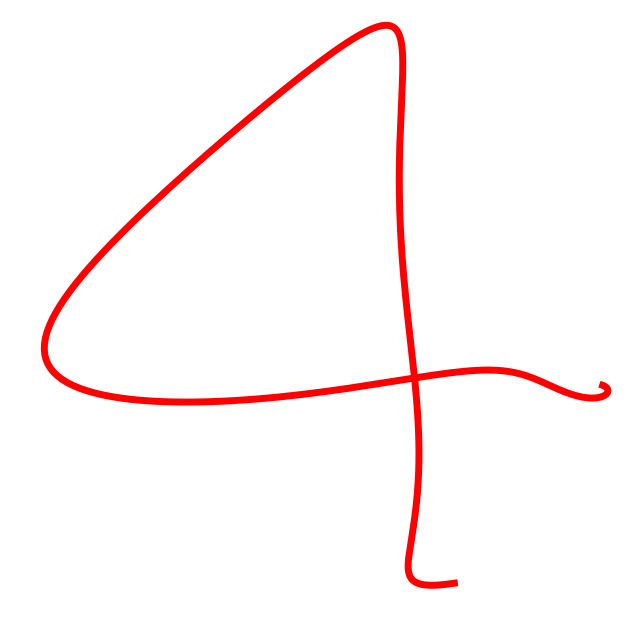}}
    \vspace{-0.2in}
    \caption*{\small Ours}
\end{subfigure}
\begin{subfigure}[b]{0.115\linewidth}
    \frame{\includegraphics[width=\linewidth]{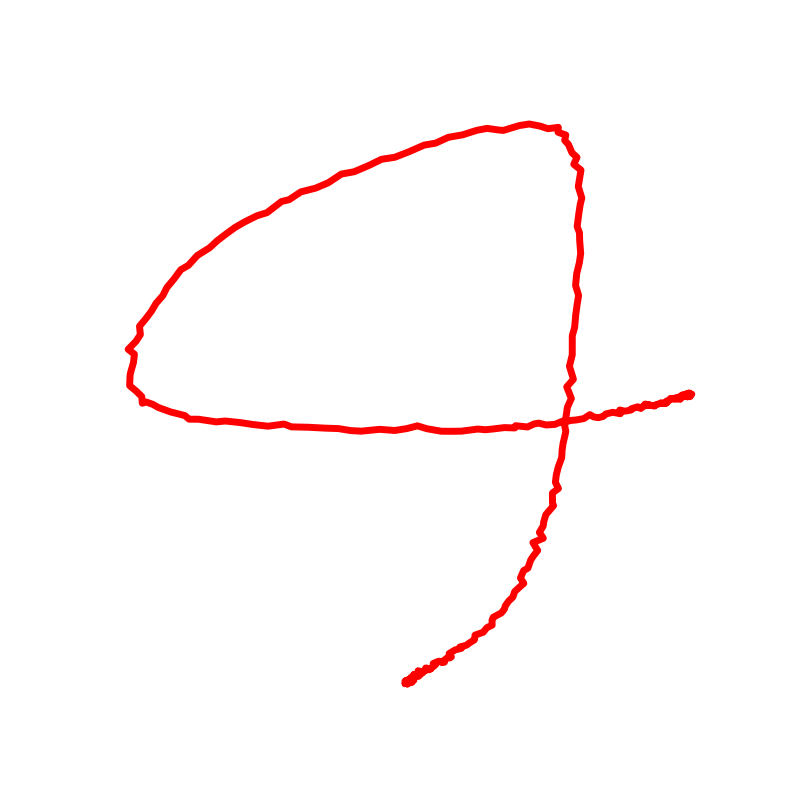}}
    \vspace{-0.2in}
    \caption*{\small CNN}
\end{subfigure}
\begin{subfigure}[b]{0.115\linewidth}
    \frame{\includegraphics[width=\linewidth]{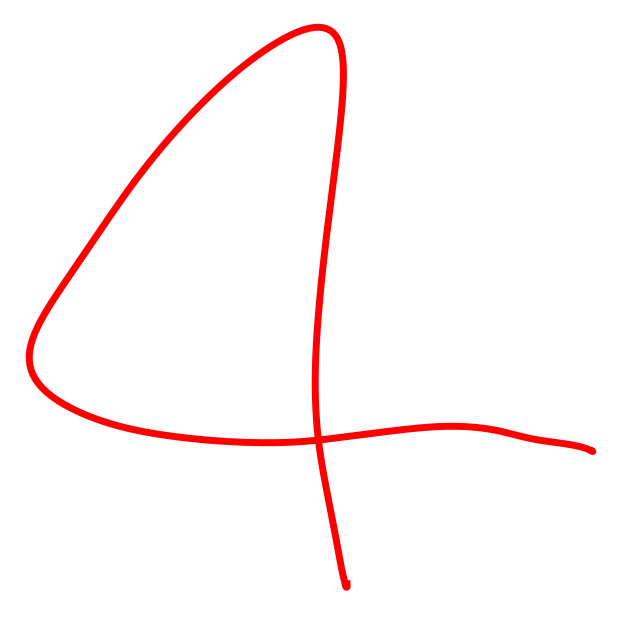}}
    \vspace{-0.2in}
    \caption*{\small CNN-DMP}
\end{subfigure}
\quad
\begin{subfigure}[b]{0.115\linewidth}
    \frame{\includegraphics[width=\linewidth]{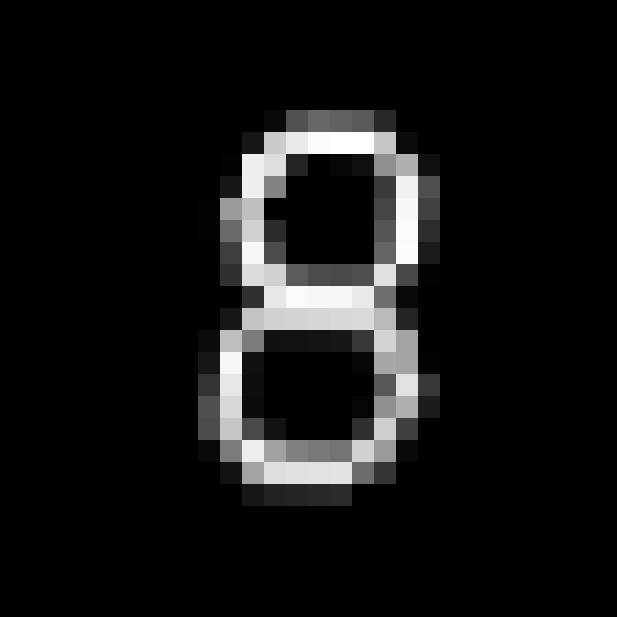}}
    \vspace{-0.2in}
    \caption*{\small Input}
\end{subfigure}
\begin{subfigure}[b]{0.115\linewidth}
    \frame{\includegraphics[width=\linewidth]{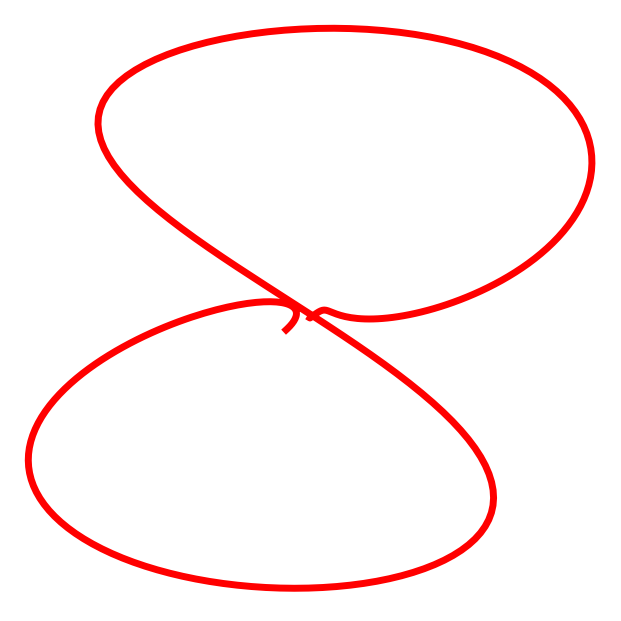}}
    \vspace{-0.2in}
    \caption*{\small Ours}
\end{subfigure}
\begin{subfigure}[b]{0.115\linewidth}
    \frame{\includegraphics[width=\linewidth]{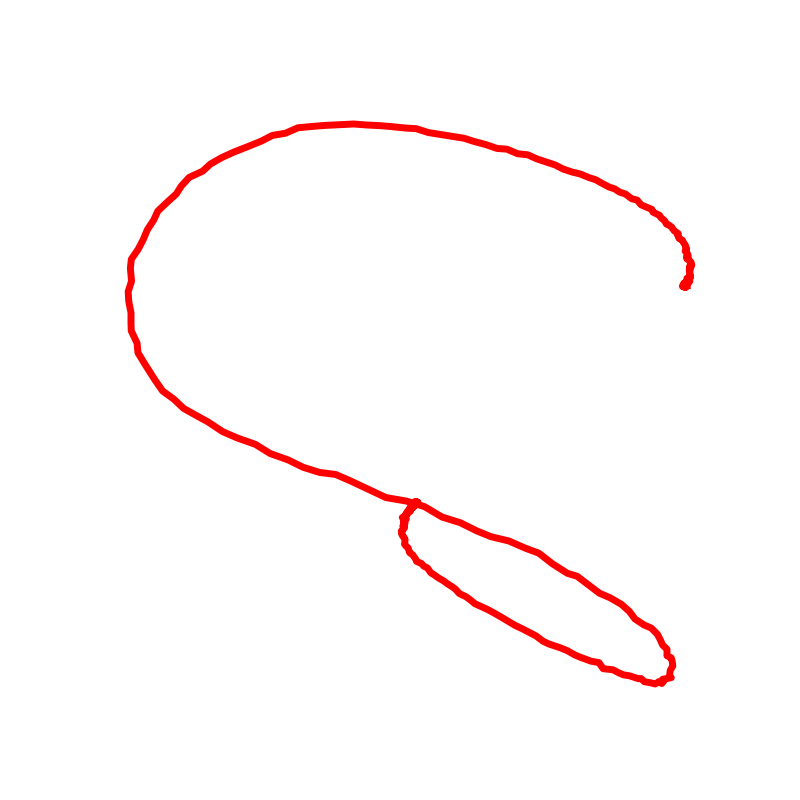}}
    \vspace{-0.2in}
    \caption*{\small CNN}
\end{subfigure}
\begin{subfigure}[b]{0.115\linewidth}
    \frame{\includegraphics[width=\linewidth]{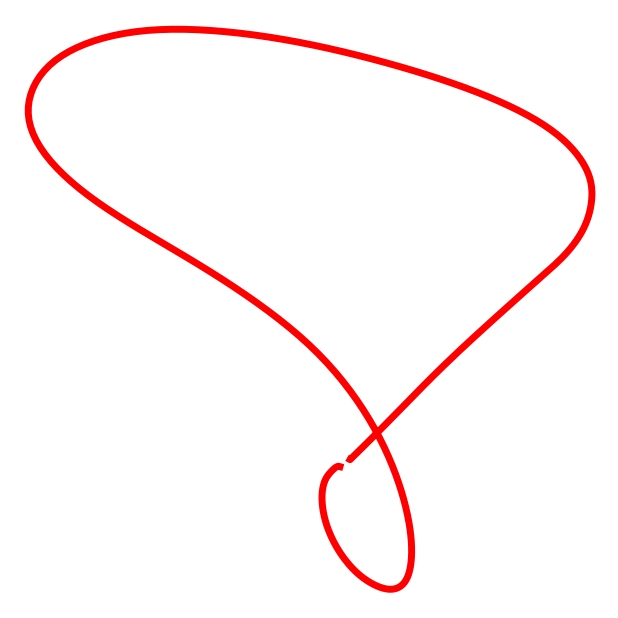}}
    \vspace{-0.2in}
    \caption*{\small CNN-DMP}
\end{subfigure}
\vspace{-0.06in}
\caption{\small Imitation (supervised) learning results on held-out test images of the digit writing task. Given an input image (left), the output action is the end-effector position of a planar robot. All methods have the same neural network architecture for fair comparison. We find that the trajectories predicted by \ours (ours) are dynamically smooth as well as more accurate than both baselines (a vanilla CNN and the architecture from \citet{pahic2018deepenc}.}
\vspace{-2mm}
\label{fig:supervised}
\end{figure}
\section{Experimental Setup}
\label{sec:baseline}

\paragraph{Environments}
\label{sec:envs}
To test our method on dynamic environments, we took existing torque control based environments for Picking and Throwing~\cite{ghosh2017divide} and modified them to enable joint angle control. The robot is a 6-DoF Kinova Jaco Arm. In Throwing, the robot tosses a cube into a bin, and in Picking, the robot picks up a cube and lifts it as high as possible. To test on quasi-static tasks, we use Pushing, Soccer, Faucet-Opening from the Meta-World~\cite{yu2019meta} task suite, as well as a setup that requires learning all 50 tasks (MT50) jointly (see Figure~\ref{fig:rl-envs}). These Meta-World environments are all in end-effector position control settings and based on a Sawyer Robot simulation in Mujoco \cite{todorov12mujoco}. In order to make the tasks more realistic, all environments have some degree of randomization. Picking and Throwing have random starting positions, while the rest have randomized goals.

\paragraph{Baselines}
We use PPO~\cite{ppo} as the underlying optimization algorithm for \ours and all the other baselines compared in the reinforcement learning setup. The first baseline is the PPO algorithm itself without the embedded dynamical structure. Further, as mentioned in the Section~\ref{sec:dmp-train}, \our is able to operate the robot at a much higher frequency than the world. Precisely, its frequency is $k$-times higher where $k$ is the \our rollout length (described in Section~\ref{sec:dmp-train}). Even though the robot moves at a higher frequency, the environment/world state is only observed at normal rate, i.e., once every $k$ robot steps and the reward computation at the intermediate $k$ steps only uses the stale environment/world state from the first one of the $k$-steps. Hence, to create a stronger baseline that can also operate at higher frequency, we create a ``PPO-multi'' baseline that predicts multiple actions and also uses our \textit{multi-action critic} architecture as described in Section~\ref{sec:dmp-rl}. All methods are compared in terms of performance measured against the environment sample states observed by the agent. In addition, we also compare to Variable Impedance Control in End-Effector Space (VICES) \cite{vices2019martin} and Dynamics-Aware Embeddings (Dyn-E) \cite{whitney2019dynamics} . VICES learns to output parameters of a PD controller or an Impedance controller directly. Dyn-E, on the other hand, using forward prediction based on environment dynamics, learns a lower dimensional action embedding.

\section{Evaluation Results: NDPs for Imitation and Reinforcement Learning}
We validate our approach on Imitation Learning and RL tasks in order to ascertain how \our compares to state-of-the-art methods. We investigate: a) Does dynamical structure in \ours help in learning from demonstrations in imitation learning setups?; b) How well do \ours perform on dynamic and quasi-static tasks in deep reinforcement learning setups compared to the baselines?; c) How sensitive is the performance of \ours to different hyper-parameter settings?

\subsection{Imitation (Supervised) Learning}
\begin{wraptable}{r}{0.42\textwidth}
\vspace{-0.22in}
\centering
\resizebox{\linewidth}{!}{%
\begin{tabular}{lcc}
\toprule
Method & NN & \textbf{NDP (ours)} \\
\midrule
Throw & 0.528 $\pm$ 0.262 & \textbf{0.642 $\pm$ 0.246} \\
Pick & \textbf{0.672 $\pm$ 0.074} & 0.408 $\pm$ 0.058 \\
Push & 0.002 $\pm$ 0.004 & \textbf{0.208 $\pm$ 0.049} \\
Soccer& 0.885 $\pm$ 0.016 & \textbf{0.890 $\pm$ 0.010} \\
Faucet & 0.532 $\pm$ 0.231 & \textbf{0.790 $\pm$ 0.059} \\
\bottomrule
\end{tabular}}
\caption{\small Imitation (supervised) learning results (success rates between 0 and 1) on Mujoco~\cite{todorov12mujoco} environments. We see that \our outperforms the neural network baseline in many tasks.}
\label{tab:supervised-robot}
\vspace{-0.14in}
\end{wraptable}

To evaluate \ours in imitation learning settings we train an agent to perform various control tasks. We evaluate \ours on the Mujoco~\cite{todorov12mujoco} environments discussed in Section~\ref{sec:envs} (Throwing, Picking, Pushing, Soccer and Faucet-Opening). Experts are trained using PPO~\cite{ppo} and are subsequently used to collect trajectories. We train an \our via the behaviour cloning  procedure described in Section~\ref{sec:dmp-sl}, on the collected expert data. We compare against a neural network policy (using roughly the same model capacity for both). Success rates in Table~\ref{tab:supervised-robot} indicate that \ours show superior performance on a wide variety of control tasks.

In order to evaluate the ability of \ours to handle complex visual data, we perform a digit-writing task using a 2D end-effector. The goal is to train a planar robot to trace the digit, given its image as input. The output action is the robot's end-effector position, and supervision is obtained by treating ground truth trajectories as demonstrations. We compare NDPs to a regular behavior cloning policy parameterized by a CNN and the prior approach which maps image to DMP parameters~\cite{pahic2018deepenc} (dubbed, CNN-DMP). CNN-DMP~\cite{pahic2018deepenc} trains a single DMP for the whole trajectory and requires supervised demonstrations. In contrast, to \ours can generate multiple DMPs across time and can be used in an RL setup as well. However, for a fair comparison, we compare both methods apples-to-apples with single DMP for whole trajectory, i.e., $k=300$.

\begin{wraptable}{r}{0.46\textwidth}
\vspace{-0.16in}
\centering
\resizebox{\linewidth}{!}{%
\begin{tabular}{lcc}
\toprule
Method & Train & Test (held-out)\\
\midrule
CNN & 10.42 $\pm$ 5.26 & 10.59 $\pm$ 4.63 \\
CNN-DMP~\cite{pahic2018deepenc} & \phantom{0}9.44 $\pm$ 4.59 & \phantom{0}8.46 $\pm$ 8.45 \\
\midrule
\our (ours) & \textbf{\phantom{0}0.70 $\pm$ 0.36} & \textbf{\phantom{0}0.74 $\pm$ 0.34} \\
\bottomrule
\end{tabular}}
\caption{\small Imitation learning on digit writing task. We report the mean loss across 10 digit classes. The input is the image of the digit to be written and action output is the end-effector position of robot. Our method significantly outperforms the baseline.}
\label{tab:supervised}
\vspace{-0.2in}
\end{wraptable}

Qualitative examples are in Figure~\ref{fig:supervised}, and quantitative results in Table~\ref{tab:supervised} report the mean loss between output trajectory and ground truth. \our outperforms both CNN and CNN-DMP \cite{pahic2018deepenc} drastically. Our method also produces much higher quality and smoother reconstructions as shown in Figure~\ref{fig:supervised}. Results show that our method can efficiently capture dynamic motions in a supervised setting, while learning from visual data.

\subsection{Reinforcement Learning}
In contrast to imitation learning where the rollout length of \our is high ($k=300$), we set $k=5$ in RL because the reward becomes too sparse if $k$ is very large. We compare the success rate of our method with that of the baseline methods PPO, PPO-multi with $k=5$, VICES and DYN-E.

\begin{figure}[t!]
\centering
\begin{subfigure}[b]{0.32\linewidth}
    \includegraphics[width=\linewidth]{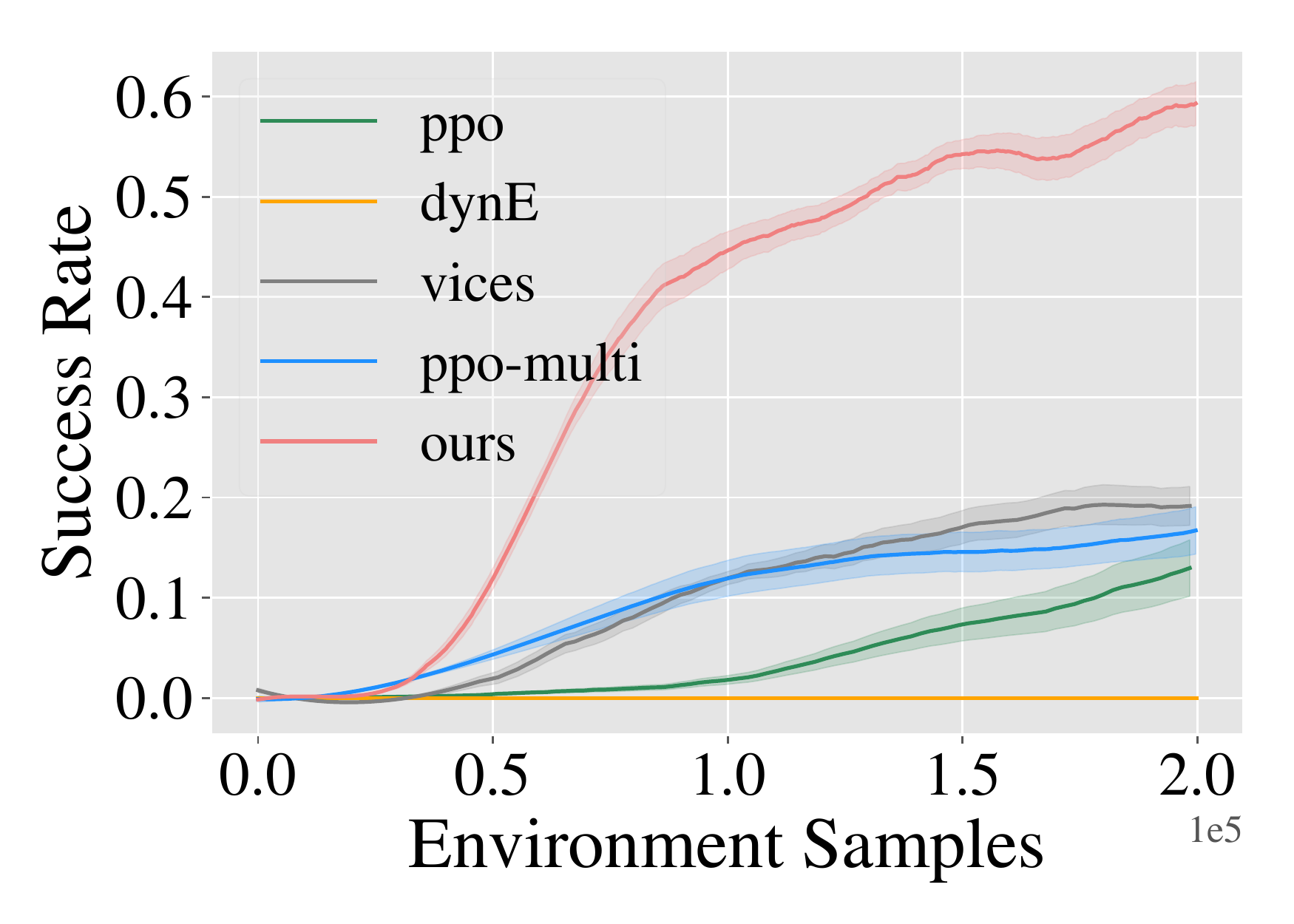}
    \vspace{-0.26in}
    \caption{\small Throwing}
    \label{fig:throw-rl}
\end{subfigure}
\begin{subfigure}[b]{0.32\linewidth}
    \includegraphics[width=\linewidth]{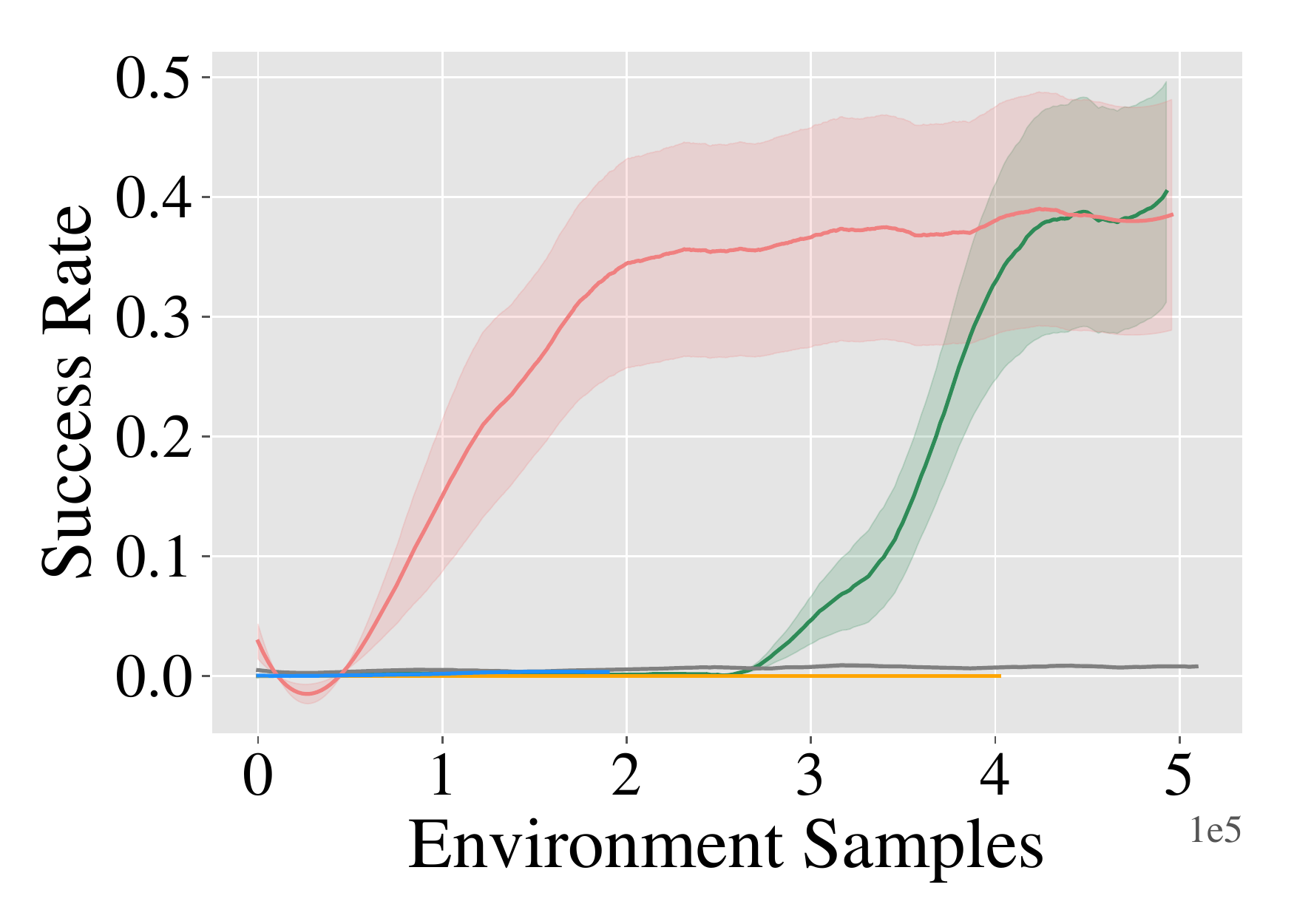}
    \vspace{-0.26in}
    \caption{\small Picking}
    \label{fig:pick-rl}
\end{subfigure}
\begin{subfigure}[b]{0.32\linewidth}
    \includegraphics[width=\linewidth]{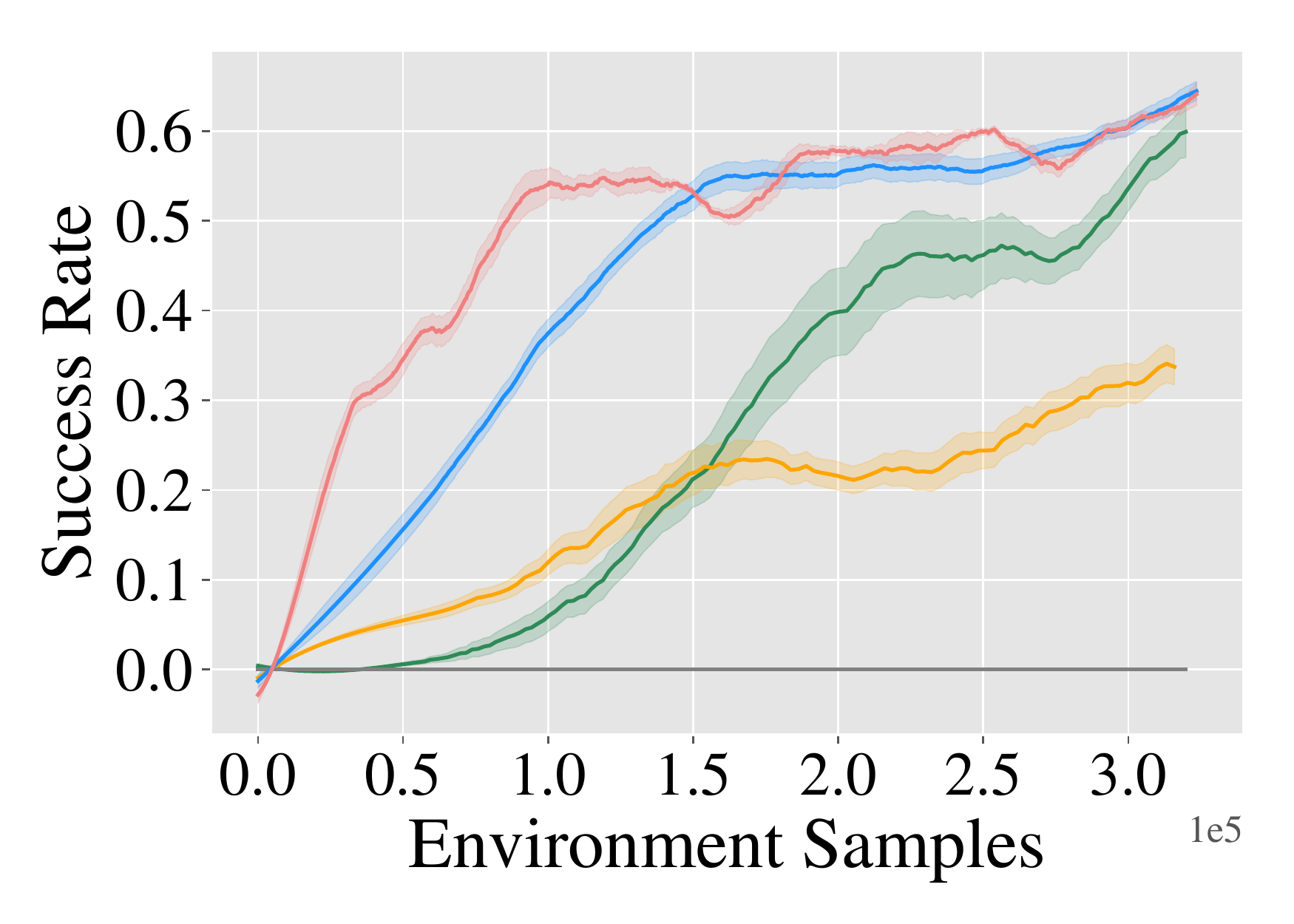}
    \vspace{-0.26in}
    \caption{\small Pushing}
    \label{fig:push-rl}
\end{subfigure}
\begin{subfigure}[b]{0.32\linewidth}
    \includegraphics[width=\linewidth]{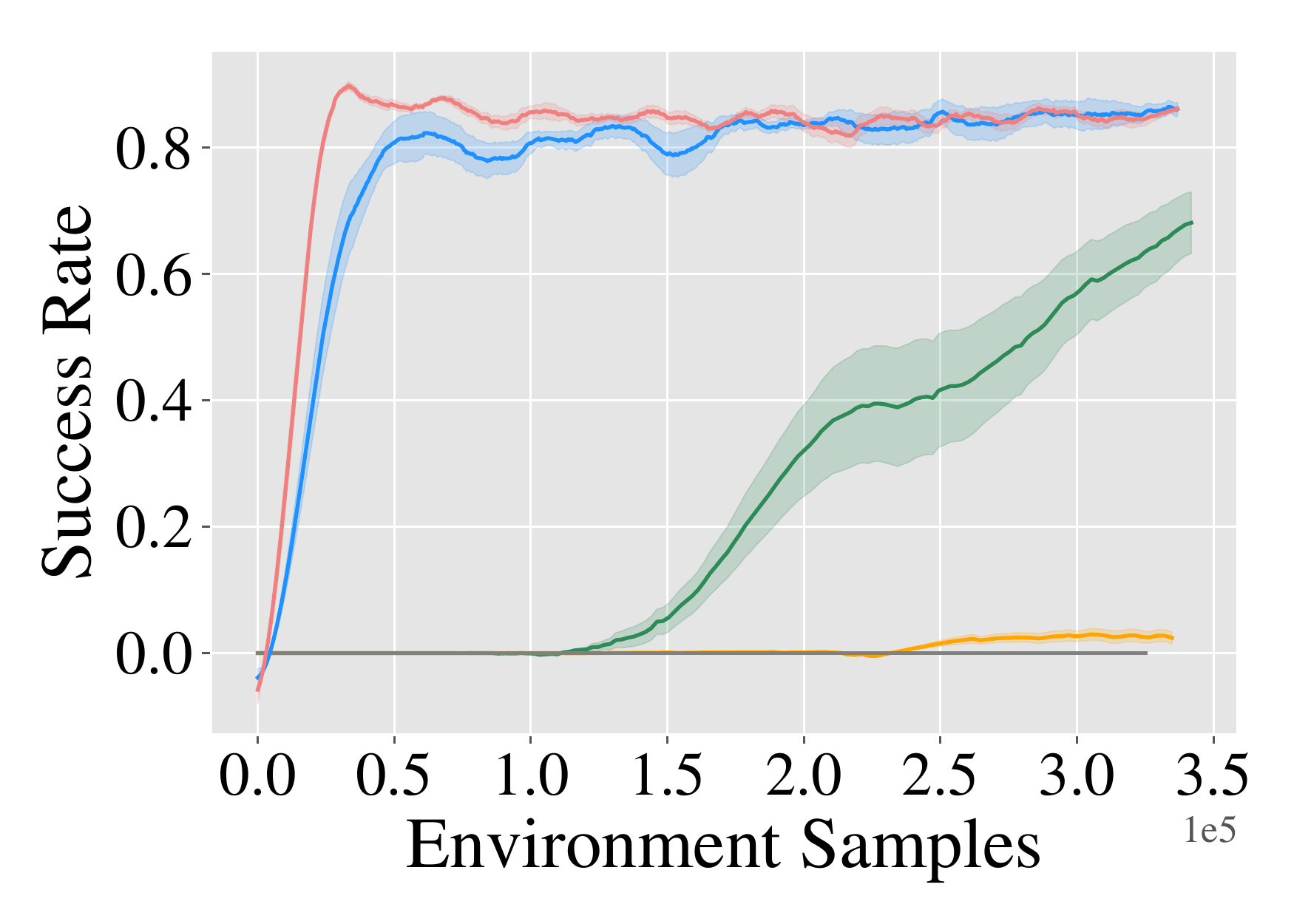}
    \vspace{-0.26in}
    \caption{\small Faucet Open}
    \label{fig:faucet-rl}
\end{subfigure}
\begin{subfigure}[b]{0.32\linewidth}
    \includegraphics[width=\linewidth]{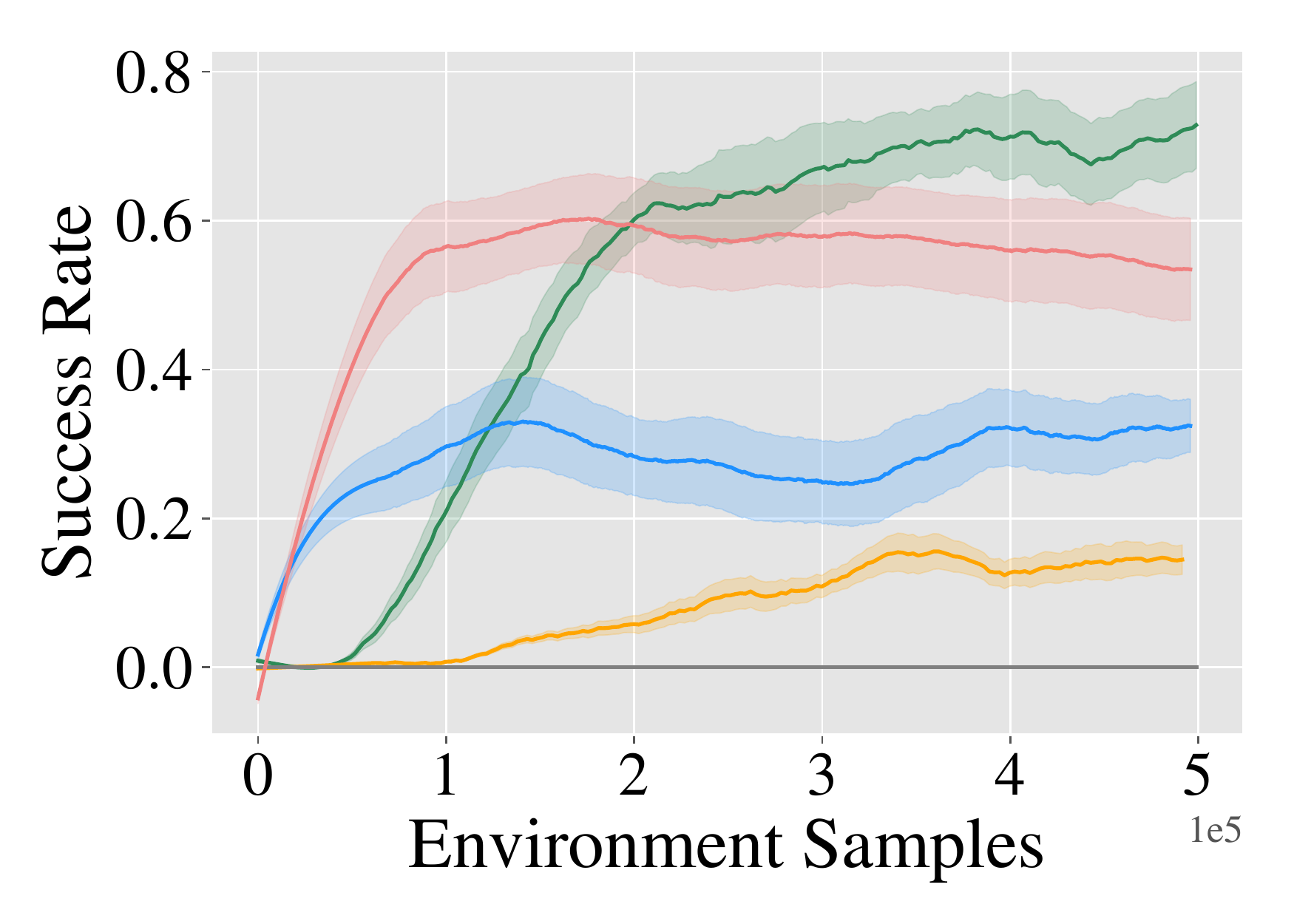}
    \vspace{-0.26in}
    \caption{\small Soccer}
    \label{fig:soccer-rl}
\end{subfigure}
\begin{subfigure}[b]{0.32\linewidth}
    \includegraphics[width=\linewidth]{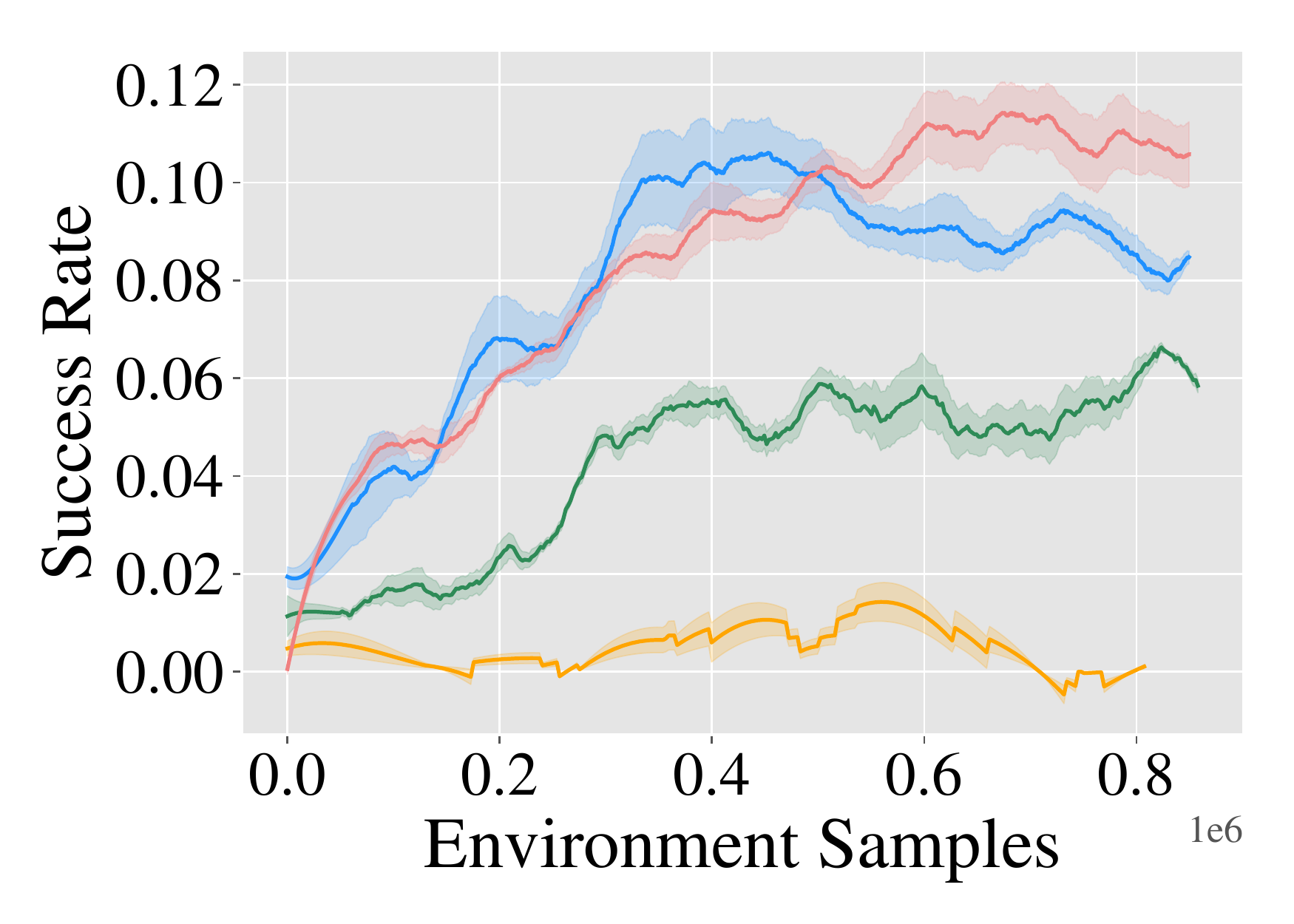}
    \vspace{-0.26in}
    \caption{\small Joint 50 MetaWorld Tasks}
    \label{fig:door-rl}
\end{subfigure}
\vspace{-0.01in}
\caption{\small Evaluation of reinforcement learning setup for continuous control tasks. Y axis is success rate and X axis is number of environment samples. We compare to PPO \cite{ppo}, a multi-action version of PPO, VICES \cite{vices2019martin} and DYN-E \cite{whitney2019dynamics}. The dynamic rollout for \our \& PPO-multi is $k=5$.}
\vspace{-0.06in}
\label{fig:rl-results}
\end{figure}

As shown in In Figure~\ref{fig:rl-results}, our method \our sees gains in both efficiency and performance in most tasks. In Soccer, PPO reaches a higher final performance, but our method shows twice the efficiency at a small loss in performance. The final task of training jointly across 50 Meta-World tasks is too hard for all methods. Nevertheless, our NDP attains a slightly higher absolute performance than baseline but doesn't show efficiency gains over baselines.

PPO-multi, a multi-action algorithm based on our proposed multi-action critic setup tends to perform well in some case (Faucet Opening, Pushing etc) but is inconsistent in its performance across all tasks and fails completely at times, (Picking etc.). Our method also outperforms prior state-of-the-art methods that re-paremeterize action spaces, namely, VICES \cite{vices2019martin} and Dyn-E \cite{whitney2019dynamics}. VICES is only slightly successful in tasks like throwing, since a PD controller can efficiently solve the task, but suffer in more complex settings due to a  large action space dimensionality (as it predicts multiple quantities per degree of freedom). Dyn-E, on the other hand, performs well on tasks such as Pushing, or Soccer, which have simpler dynamics and contacts, but fails to scale to more complex environments.

Through these experiments, we show the diversity and versatility of \our, as it has a strong performance across different types of control tasks. \our outperforms baselines in both dynamic (throwing) and static tasks (pushing) while being able to learn in a more data efficient manner. It is able to reason in a space of physically meaningful trajectories, but it does not lose the advantages and flexibility that other policy setups have.

\begin{figure}[t!]
\centering
\begin{subfigure}[b]{0.24\linewidth}
    \includegraphics[width=\linewidth]{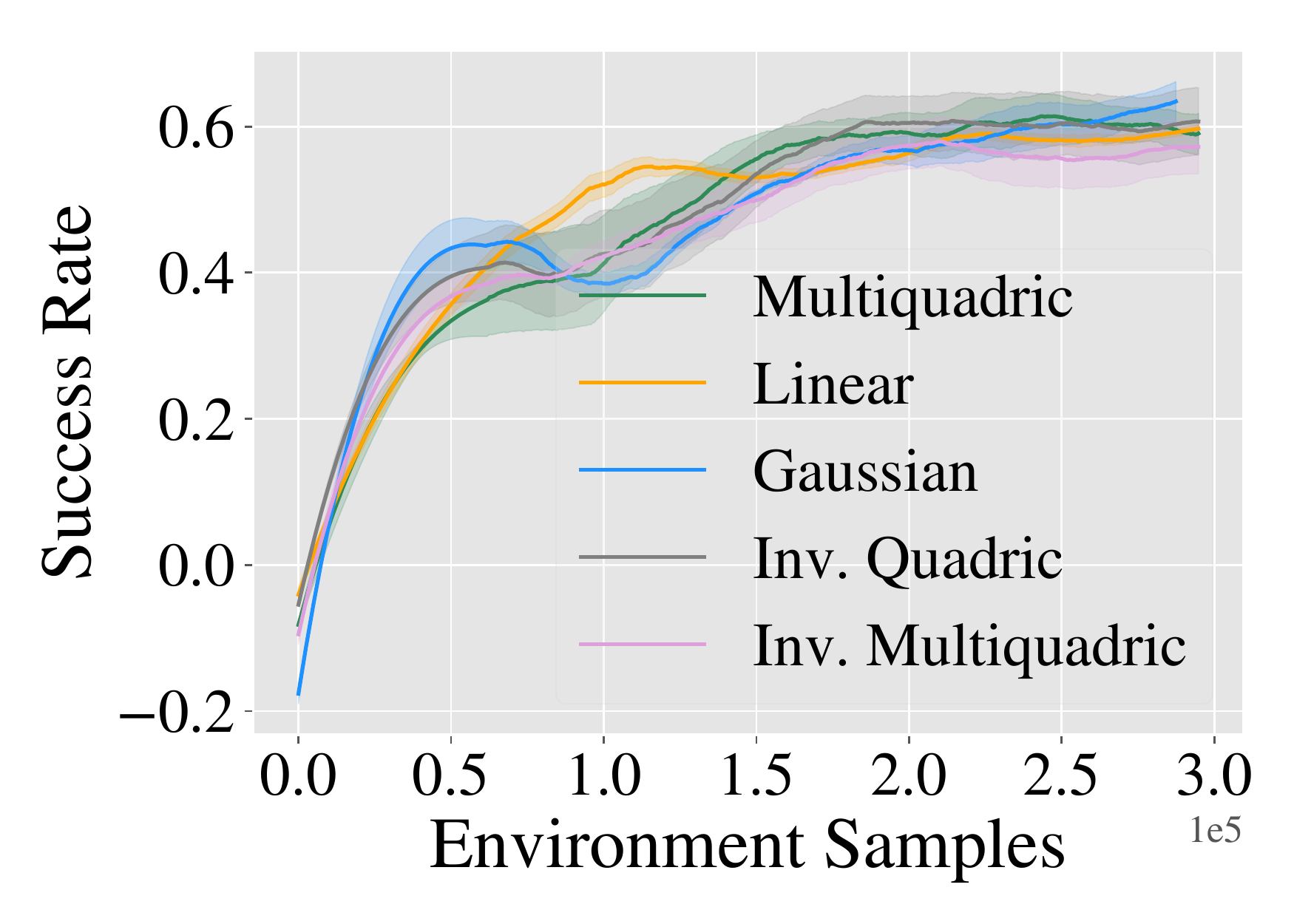}
    \vspace{-0.2in}
    \caption{\small RBF Kernels}
    \label{fig:rl-ablation-1}
\end{subfigure}
\begin{subfigure}[b]{0.24\linewidth}
    \includegraphics[width=\linewidth]{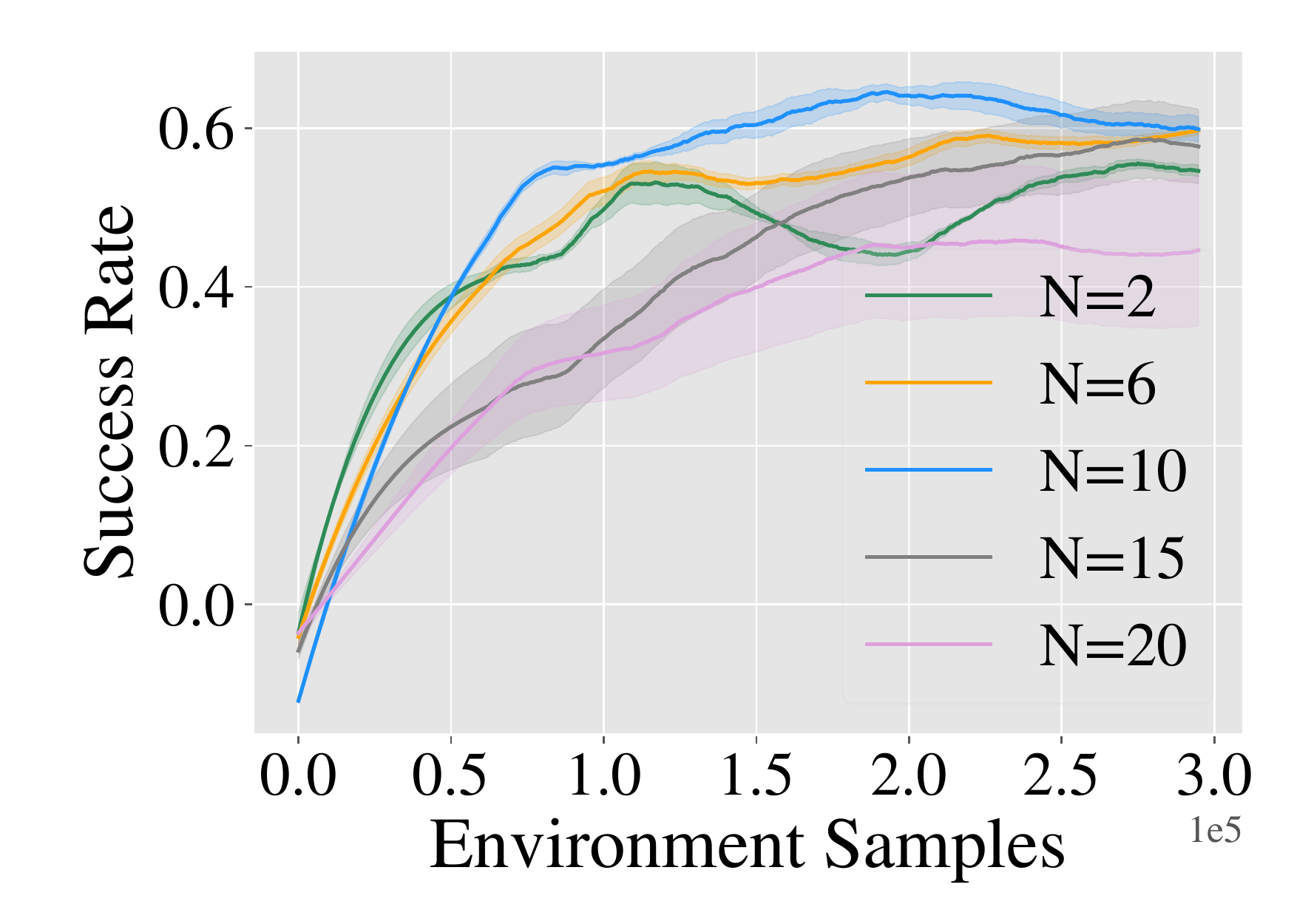}
    \vspace{-0.2in}
    \caption{\small \# of basis functions}
    \label{fig:rl-ablation-2}
\end{subfigure}
\begin{subfigure}[b]{0.24\linewidth}
    \includegraphics[width=\linewidth]{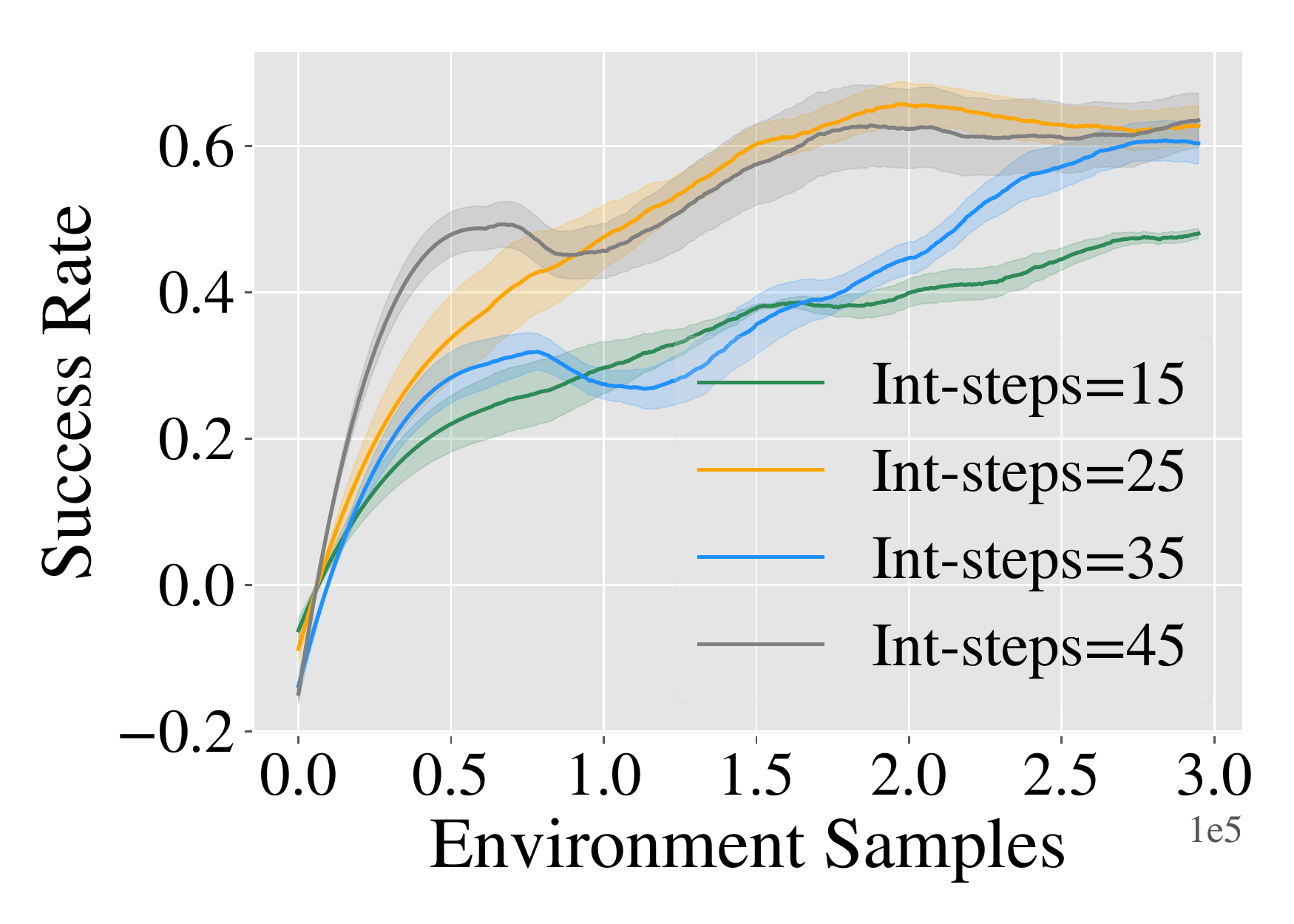}
    \vspace{-0.2in}
    \caption{\small Integration steps}
    \label{fig:rl-ablation-3}
\end{subfigure}
\begin{subfigure}[b]{0.24\linewidth}
    \includegraphics[width=\linewidth]{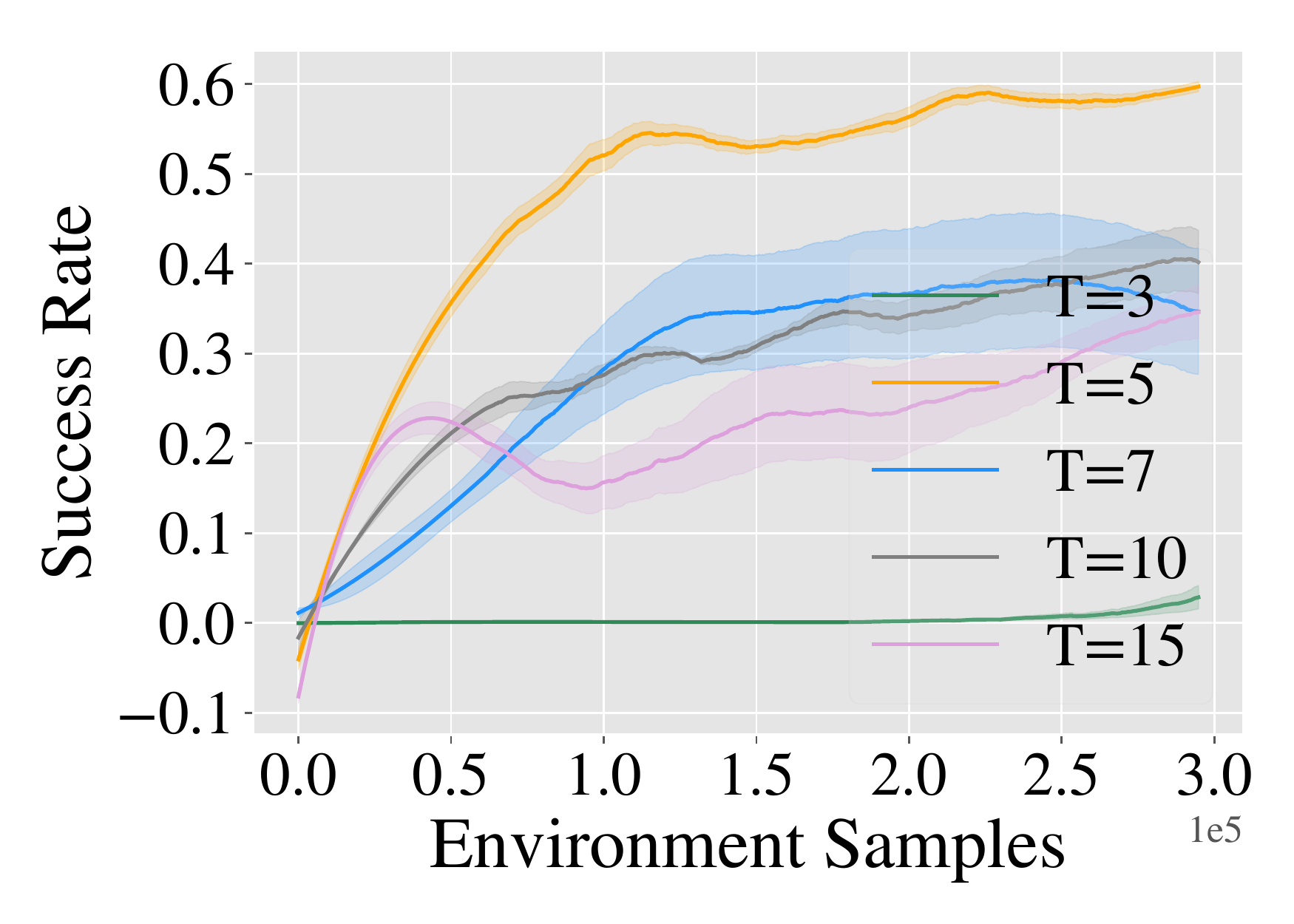}
    \vspace{-0.2in}
    \caption{\small Rollout length}
    \label{fig:rl-ablation-4}
\end{subfigure}
\vspace{-0.05in}
\caption{\small Ablation of \ours with respect to different hyperparameters in  the RL setup (pushing). We ablate different choices of radial basis functions in (a). We ablate across number of basis functions, integration steps, and length of the \our rollout in (b,c,d). Plots indicate that \ours are fairly stable across a wide range of choices.}
\vspace{-0.08in}
\label{fig:rl-ablations}
\end{figure}

\subsubsection{Ablations for NDPs in Reinforcement Learning Setup}
We aim to understand how design choices affect the RL performance of \our. We run comparisons on the pushing task, varying the number of basis functions $N$ (in the set $\{2, 6, 10, 15, 20\}$), DMP rollout lengths (in set $\{3, 5, 7, 10, 15\}$), number of integration steps (in set $\{15, 25, 35, 45\}$), as well as different basis functions: Gaussian RBF (standard), $\psi$ defined in Equation~\eqref{eq:forcing_func}, a liner map $\psi(x) = x$, a multiquadric map: $\psi(x) = \sqrt{1 + (\epsilon x)^2}$, a inverse quadric map $\psi(x) = \frac{1}{1 + (\epsilon x)^2}$, and an inverse multiquadric map: $\psi(x) = \frac{1}{\sqrt{1 + (\epsilon x)^2}}$.

Additionally, we investigate the effect of different \our components on its performance. To this end, we ablate a setting where only $g$ (the goal) is learnt while the radial basis function weights (the forcing function) are $0$ (we call this setting `only-g'). We also ablate a version of \our that learns the global constant $\alpha $ (from Equation~\ref{eq:accel_dmp_diff}), in addition to the other parameters ($g$ and $w$).

\begin{figure}[h!]
\centering
\begin{subfigure}[b]{0.27\linewidth}
    \includegraphics[width=\linewidth]{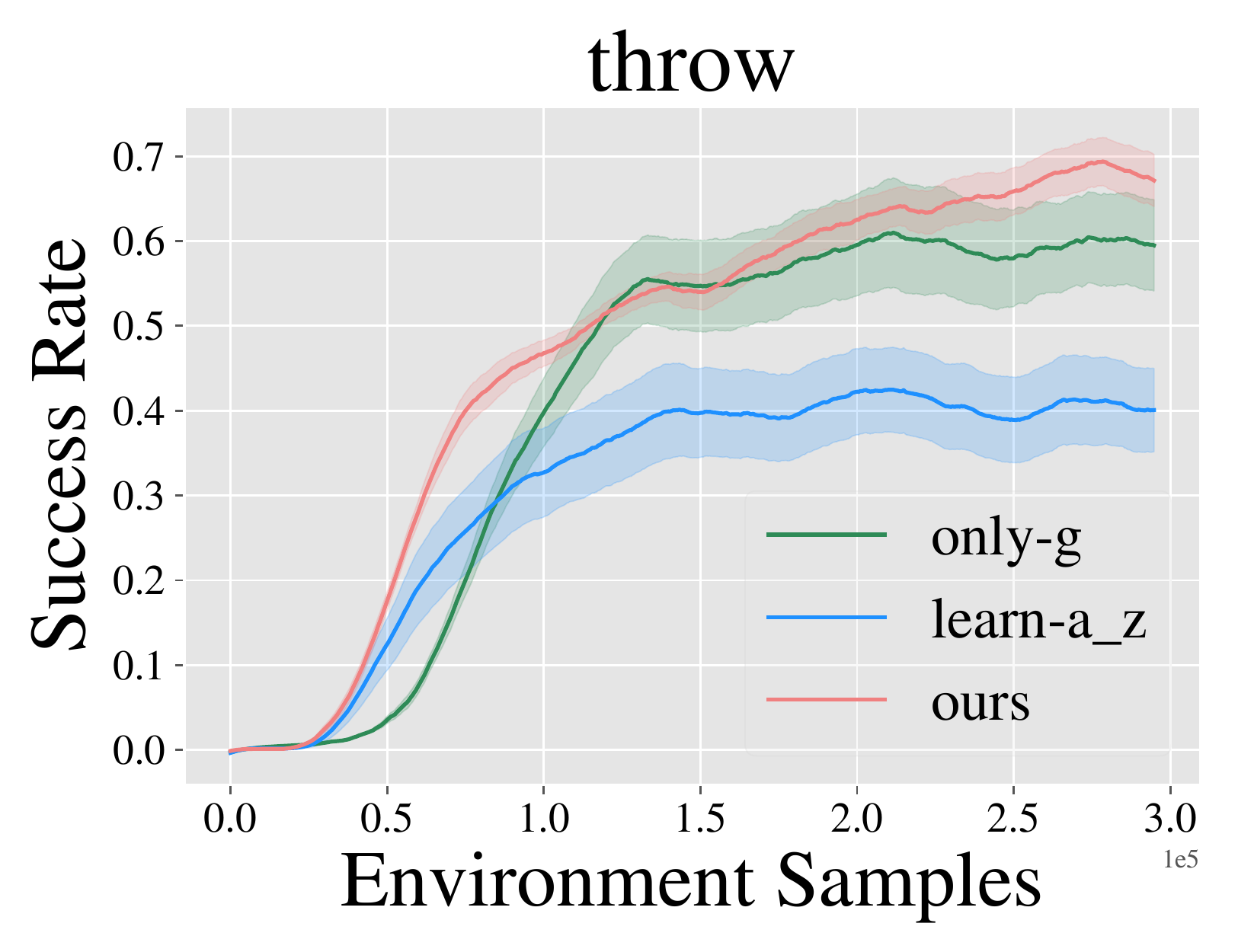}
    \vspace{-0.26in}
    \caption{\small Throwing}
    \label{fig:throw-ablation-2}
\end{subfigure}
\begin{subfigure}[b]{0.27\linewidth}
    \includegraphics[width=\linewidth]{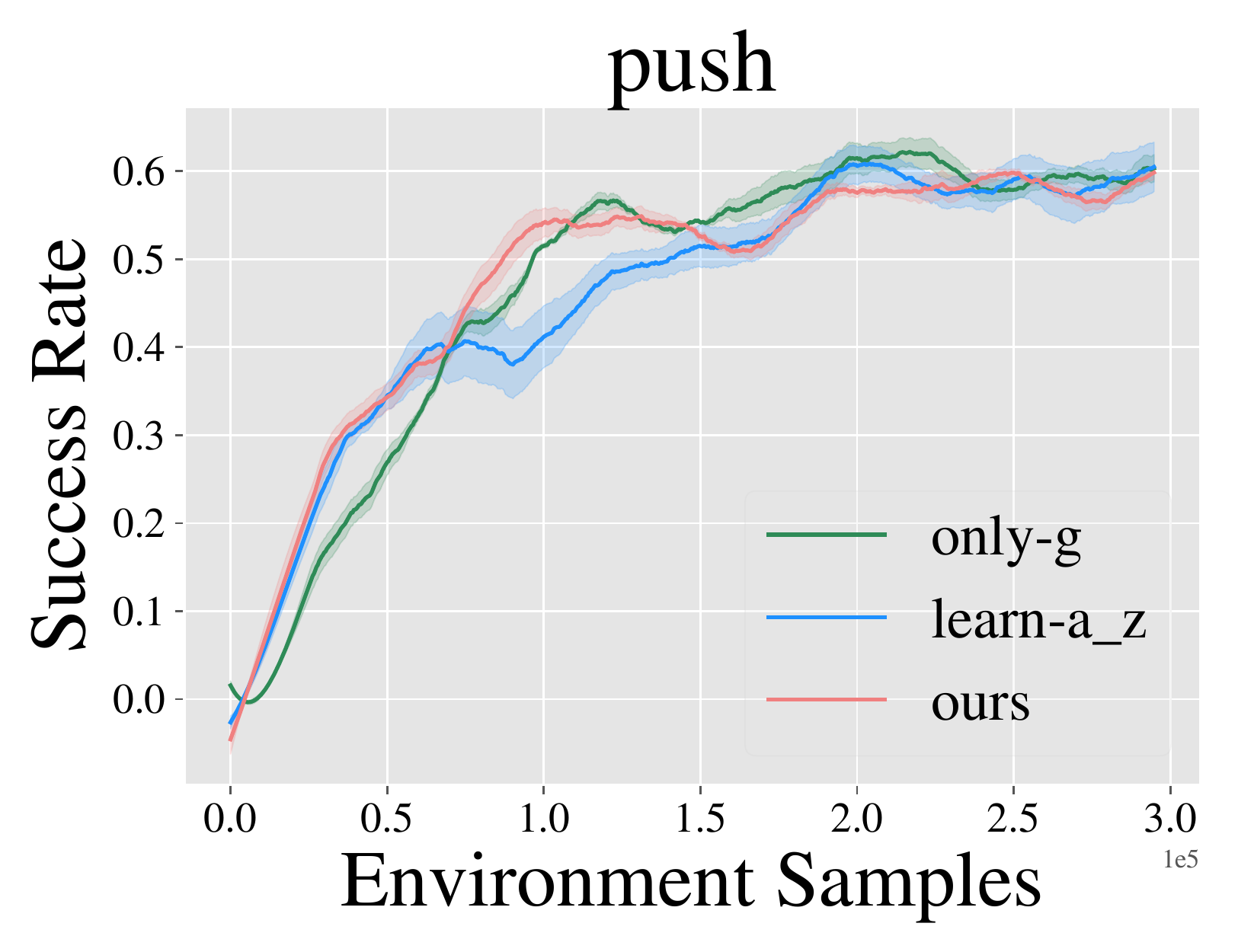}
    \vspace{-0.26in}
    \caption{\small Push}
    \label{fig:push-ablation-2}
\end{subfigure}
\begin{subfigure}[b]{0.27\linewidth}
    \includegraphics[width=\linewidth]{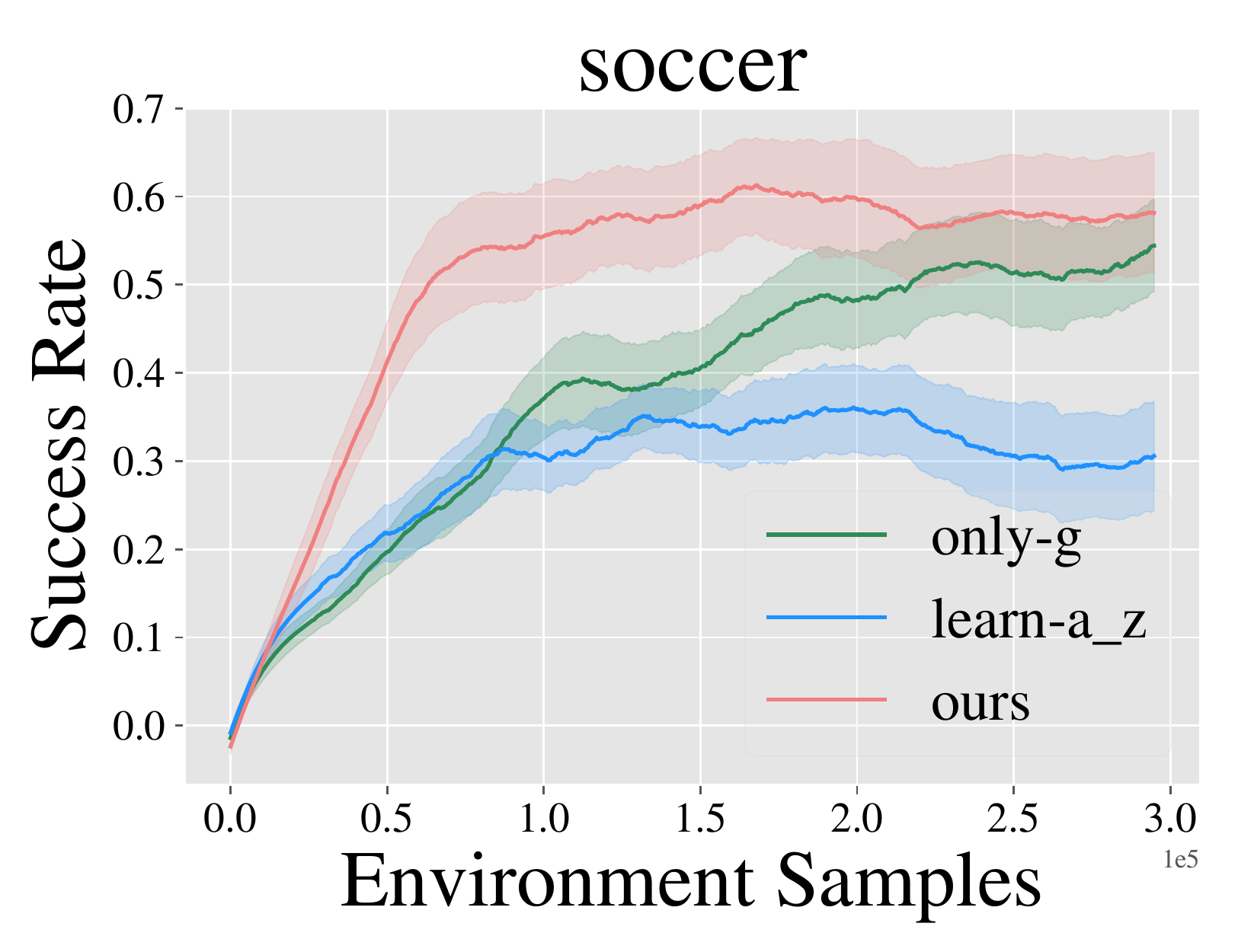}
    \vspace{-0.26in}
    \caption{\small Soccer}
    \label{fig:soccer-ablation-2}
\end{subfigure}
\begin{subfigure}[b]{0.27\linewidth}
    \includegraphics[width=\linewidth]{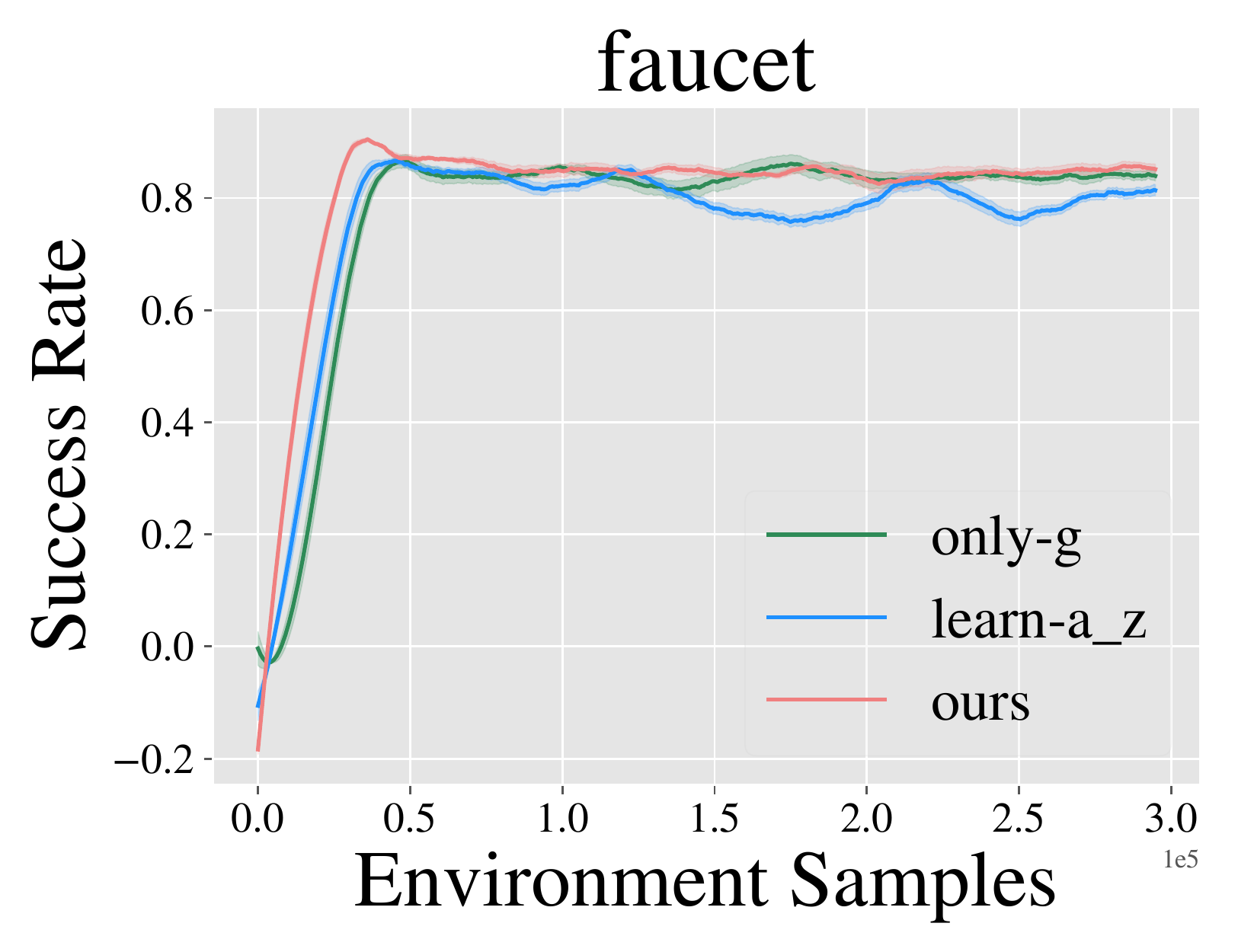}
    \vspace{-0.26in}
    \caption{\small Faucet Open}
    \label{fig:faucet-ablation-2}
\end{subfigure}
\begin{subfigure}[b]{0.27\linewidth}
    \includegraphics[width=\linewidth]{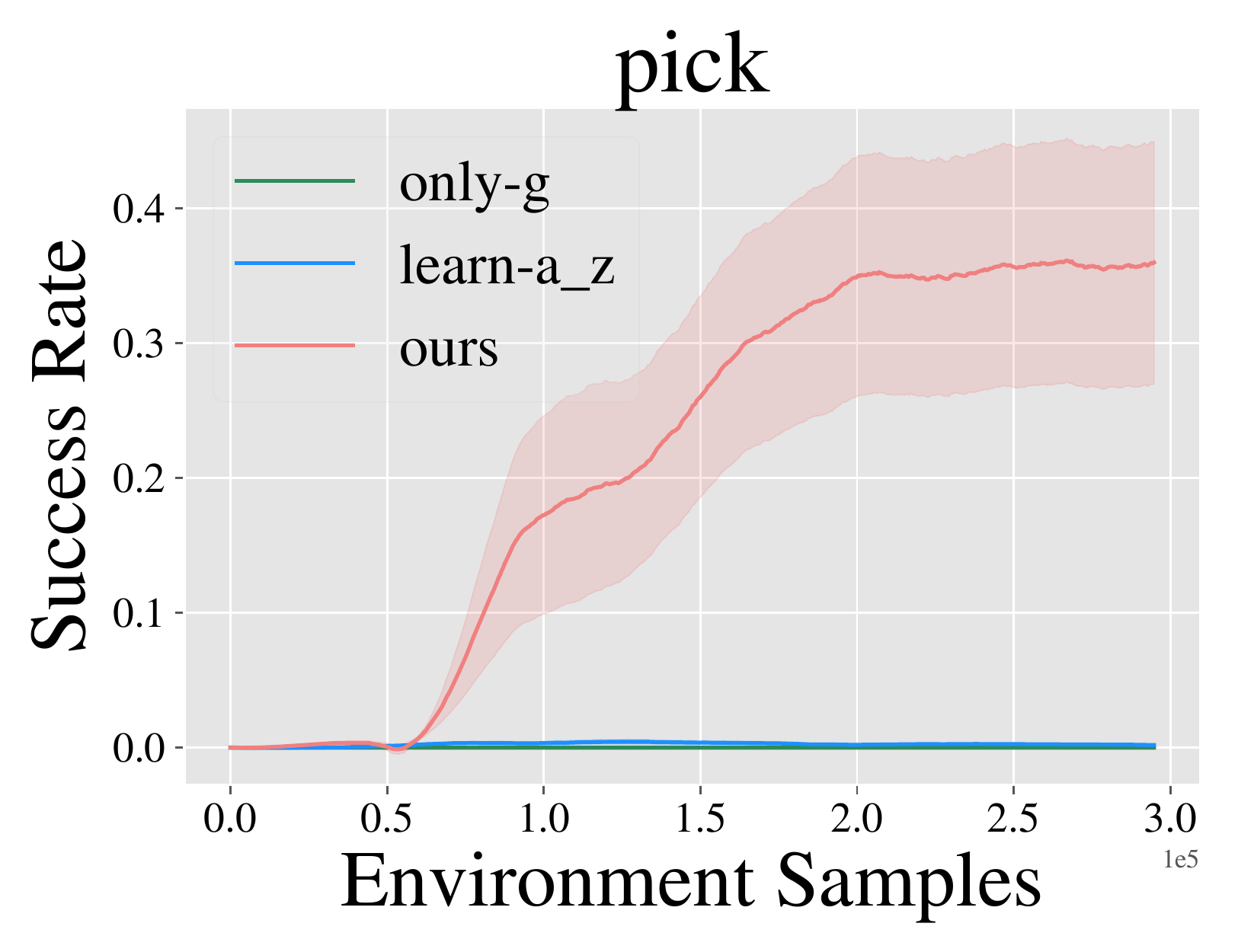}
    \vspace{-0.26in}
    \caption{\small Picking}
    \label{fig:pick-ablation-2}
\end{subfigure}
\vspace{-0.01in}
\caption{\small Ablations for different \our design choices. The first entails \our also learning the parameter $\alpha$ (shown as $a_z$). In the second one, $g$ is learnt but not $w_i$, i.e. the forcing function is $0$ (`only-g'). Results indicate that \our outperforms both these settings.}
\vspace{-0.06in}
\label{fig:ablation-2}
\end{figure}

Figure~\ref{fig:rl-ablations} shows results from ablating different \our parameters. Varying $N$ (number of basis functions) controls the shape of the trajectory taken by the agent. A small $N$ may not have the power to represent the nuances of the motion required, while a big $N$ may make the parameter space too large to learn efficiently. We see that number of integration steps do not have a large effect on performance, similarly to the type of radial basis function. Most radial basis functions generally have similar interpolation and representation abilities. We see that $k=3$ (the length of each individual rollout within \our) has a much lower performance due to the fact that 3 steps cannot capture the smoothness or intricacies of a trajectory. Overall, we mostly find that \our is robust to design choices.  Figure~\ref{fig:ablation-2} shows that the current formulation of \our outperforms the one where $\alpha$ is learnt. We also observe that setting the forcing term to 0 (only learning the goal, $g$) is significantly less sample efficient than \ours while converging to a slightly lower asymptotic performance.

\section{Related Work}
\paragraph{Dynamic Movement Primitives} Previous works have proposed and used Dynamic Movement Primitives (DMP) \cite{schaal2006dynamic,isprt2012dmp,prada2013dmp} for robot control~\cite{mulling2013learning, peters2003reinforcement, kormushev2010robot}.
Work has been done in representing dynamical systems, both as extensions of DMPs \cite{conkey2019promp,Calinon2010LearningbasedCS,Calinon2016ATO,ude2010task}, and beyond DMPs by learning kernels \cite{huang2019kmp} and designing Riemannian metrics \cite{ratliff2018riemannian}.
Learning methods have been used to incorporate environment sensory information into the forcing function of DMPs~\cite{sutanto2018learning,rai2017learning}.
DMPs have also been prime candidates to represent primitives when learning hierarchical policies, given the range of motions DMPs can be used for in robotics \cite{daniel2016hreps, stulp2012sequences, kober2009learning, pastor2011skill}. Parametrization of DMPs using gaussian processes has also been proposed to facilitate generalization~\cite{ude2010task,pervez2018learning}. Recently, deep learning has also been used to map images to DMPs~\cite{pahic2018deepenc} and to learn DMPs in latent feature space~\cite{chen2016dynamic}. However most of these works require pre-trained DMPs via expert demonstrations or are only evaluated in the supervised setting. Furthermore, either a single DMP is used to represent the whole  trajectory or the demonstration is manually segmented to learn a different DMP for each segment. In contrast, \ours outputs a new dynamical system for each timestep to fit diverse trajectory behaviours across time. Since we embed dynamical structure into the deep network, \ours can flexibly be incorporated not just in visual imitation but also in deep reinforcement learning setups, in an end-to-end manner.

\paragraph{Reparameterized Policy and Action Spaces}
A broader area of work that makes use of action reparameterization is the study of Hierarchical Reinforcement Learning (HRL). Works in the options framework \cite{bacon2017option, sutton1999temporal} attempt to learn an overarching policy that controls usage of lower-level policies or primitives. Lower-level policies are usually pre-trained therefore require supervision and knowledge of the task beforehand, limiting the generalizability of such methods. For example, \citet{daniel2016hreps, parisi2015tetherball} incorporate DMPs into option-based RL policies, using a pre-trained DMPs as the high level actions. This setup requires re-learning DMPs for different types of tasks and does not allow the same policy to generalize, since it needs to have access to an extremely large number of DMPs. Action spaces can also be reparameterized in terms of pre-determined PD controller~\cite{yuan2019ego} or learned impedance controller parameters~\cite{vices2019martin}. While this helps for policies to adapt to contact rich behaviors, it does not change the trajectories taken by the robot. This often leads to high dimensionality, and thus a decrease sample efficiency. In addition, \citet{whitney2019dynamics} learn an action embedding based on passive data, however, this does not take environment dynamics or explicit control structure into account.

\paragraph{Structure in Policy Learning}
Various methods in the field of control and robotics have employed physical knowledge, dynamical systems, optimization, and more general task/environment dynamics to create structured learning. Works such as \cite{cranmer2020lagrangian, greydanus2019hamiltonian} propose networks constrained through physical properties such as Hamiltonian co-ordinates or Lagrangian Dynamics. However, the scope of these works is limited to toy examples such as a point mass, and are often used for supervised learning. Similarly, other works \cite{rana2020euclideanizing, ravichandar2017demonstration, perrin2016fastdiffeomorphic, neumann2015diffeomorphic} all employ dynamical systems to model demonstrations, and do not tackle generalization or learning beyond imitation. Fully differentiable optimization problems have also been incorporated as layers inside a deep learning setup \cite{amos2017optnet,chen2018neural,amos2019diffmpc}. Whil they share the underlying idea of embedding structure in deep networks such that some aspects of this structure can be learned end-to-end, they have not been explored in tackling complex robotic control tasks. Furthermore, it is common in RL setups to incorporate planning based on a system model \cite{deisenroth2011pilco, Deisenroth_2015, chua2018deep, Atkeson97acomparison, deisenroth2013policy}. However, this is usually learned from experience or from attempts to predict the effects of actions on the environment (forward and inverse models), and often tends to fail for complex dynamic tasks.

\section{Discussion}
Our method attempts to bridge the gap between classical robotics, control and recent approaches in deep learning and deep RL. We propose a novel re-parameterization of action spaces via Neural Dynamic Policies, a set of policies which impose the structure of a dynamical system on action spaces. We show how this set of policies can be useful for continuous control with RL, as well as in supervised learning settings. Our method obtains superior results due to its natural imposition of structure and yet it is still generalizable to almost any continuous control environment.

The use of DMPs in this work was a particular design choice within our architecture which allows for any form of dynamical structure that is differentiable. As alluded to in the introduction, other similar representations can be employed in their place. In fact, DMPs are a special case of a general second order dynamical system~\cite{bullo2005geometric,ratliff2018riemannian} where the inertia term is identity, and potential and damping functions are defined in a particular manner via first order differential equations with a separate forcing function which captures the complexities of the desired behavior. Given this, one can setup a dynamical structure such that it explicitly models and learns the metric, potential, and damping explicitly. While this brings advantages in better representation, it also brings challenges in learning. We leave these directions for future work to explore.

\section*{Acknowledgments}
We thank Giovanni Sutanto, Stas Tiomkin and Adithya Murali for fruitful discussions. We also thank Franziska Meier, Akshara Rai, David Held, Mengtian Li, George Cazenavette, and Wen-Hsuan Chu for comments on early drafts of this paper. This work was supported in part by DARPA Machine Common Sense grant and Google Faculty Award to DP.

\bibliographystyle{abbrvnat}
\bibliography{main}
\clearpage

\appendix
\section{Appendix}

\subsection{Videos}
Videos can also be found at: \url{https://shikharbahl.github.io/neural-dynamic-policies/}.  We found that NDP results look dynamically more stable and smooth in comparison to the baselines. PPO-multi generates shaky trajectories, while corresponding NDP (ours) trajectories are much smoother .This is perhaps due to the embedded dynamical structure in NDPs as all other aspects in PPO-multi and NDP (ours) are compared apples-to-apples.

\subsection{Differentiability Proof of Dynamical Structure in NDPs}
In Section 3.2, we provide an intuition for how \our is incorporates a second order dynamical system (based on the DMP system, described in Section 2) in a differentiable manner. Let us start by observing that, when implementing our algorithm, $y_0$, $\dot{y}_0$ are known and $\ddot{y}_0 = 0$, as well as $x_0 = 1$. Assuming that the output states of \our are $y_0, y_1, ..., y_t, ...$ and assuming that there exists a loss function $L$ which takes in $y_t$, we want partial derivatives with respect to DMP weights $w_i$ and goal $g$: \\  
\begin{equation}
    \frac{\partial{L(y_t)}}{\partial{w_i}} , \quad \frac{\partial{L(y_t)}}{\partial{w_i}}
\end{equation}
\begin{equation}
    \frac{\partial{L(y_t)}}{\partial{y_t}}\frac{\partial{y_t}}{\partial{w_i}}
\end{equation}

Starting with $w_i$, using the Chain Rule we get that
\begin{equation}
    \frac{\partial{L(y_t)}}{\partial{w_i}} =  \frac{\partial{L(y_t)}}{\partial{y_t}}\frac{\partial{y_t}}{\partial{w_i}}
\end{equation}

Hence, we want to be able to calculated $\frac{\partial{y_t}}{\partial{w_i}}$. For simplicity let: 
\begin{equation}
    W_t = \frac{\partial{y_t}}{\partial{w_i}}
\end{equation}
\begin{equation}
    \dot{W}_t = \frac{\partial{\dot{y}_t}}{\partial{w_i}}
\end{equation}
\begin{equation}
    \ddot{W}_t = \frac{\partial{\ddot{y}_t}}{\partial{w_i}}
\end{equation}

From section 3.2 we know that: 
\begin{equation}
\ddot{y}_t = \alpha(\beta(g - y_{t - 1}) - \dot{y}_{t - 1} + f(x_t, g)
\label{eq:accel_dmp_diff_2}
\end{equation}
and, the discretization over a small time interval $dt$ gives us: 
\begin{equation}
\dot{y}_t = \dot{y}_{t - 1} + \ddot{y}_{t - 1}dt, \quad y_t = y_{t - 1} + \dot{y}_{t - 1}dt
\label{eq:dmp_diff_2}
\end{equation}

From these and the fact that $y_0$, $\dot{y}_0$ are known and $\ddot{y}_0 = 0$, as well as $x_0 = 1$, we get that $y_1 = y_0 + \dot{y}_0dt$ and $\dot{y}_1 = \dot{y}_0 + 0 dt = \dot{y}_0$, as well as $\ddot{y}_1 =  \alpha(\beta(g - y_{0}) - \dot{y}_{0} + f(x_1, g)$. 

Using Equations~\eqref{eq:accel_dmp_diff_2} and \eqref{eq:dmp_diff_2} we get that: 
\begin{equation}
    W_t = \frac{\partial}{\partial{w_i}}( y_{t - 1} + \dot{y}_{t - 1}dt)
\end{equation}
\begin{equation}
    W_t = W_{t - 1} + \dot{W}_{t - 1}dt
    \label{eq:deriv}
\end{equation}
and 
\begin{equation}
    \dot{W}_{t - 1} = \dot{W}_{t - 2} + \ddot{W}_{t - 1} dt
    \label{eq:deriv_dot}
\end{equation}

In turn, 
\begin{equation}
    \ddot{W}_{t - 1} =   \frac{\partial}{\partial{w_i}}( \alpha(\beta(g - y_{t - 2}) - \dot{y}_{t - 2}) + f(x_{t-1}, g))  
\end{equation}

From section 3.2, we know that 
\begin{equation}
    \frac{\partial{f(x_{t-1}, g)}}{\partial{w_i}} = \frac{\psi_i}{\sum_j \psi_j}(g - y_0)x_{t-1}
\end{equation}

Hence: 
\begin{equation}
    \ddot{W}_{t - 1} =   \alpha(\beta(- W_{t - 2}) - \dot{W}_{t - 2}) +   \frac{\psi_i}{\sum_j \psi_j}(g - y_0)x_{t-1}
\end{equation}

Plugging equations the value of $\ddot{W}_{t - 1}$ into Equation~\eqref{eq:deriv_dot}: 
\begin{equation}
     \dot{W}_{t - 1} = \dot{W}_{t - 2} + (\alpha(\beta(- W_{t - 2}) - \dot{W}_{t - 2}) +   \frac{\psi_i}{\sum_j \psi_j}(g - y_0)x_{t-1}) dt
\end{equation}

Now plugging the value of $\dot{W}_{t - 1}$ in Equation~\eqref{eq:deriv}: 
\begin{equation}
    W_t = W_{t - 1} + (\dot{W}_{t - 2} + (\alpha(\beta(- W_{t - 2}) - \dot{W}_{t - 2}) +   \frac{\psi_i}{\sum_j \psi_j}(g - y_0)x_{t-1}) dt)dt
    \label{eq:final_diff}
\end{equation}

We see that the value of $W_t$ is dependent on $W_{t - 1}, \dot{W}_{t - 2},  W_{t - 2}$. We can now show that we can acquire a numerical value for $W_t$ by recursively following the gradients, given that $W_{t - 1}, \dot{W}_{t - 2},  W_{t - 2}$ are known. Since we showed that $y_0$, $\dot{y}_0$, $y_1$,$\dot{y}_1$ do not require $w_i$ in their computation, $W_1, \dot{W}_0,  W_0 = 0$. Hence by recursively following the relationship defined in Equation~\eqref{eq:final_diff}, we achieve a solution for $W_t$. 

Similarly, let: 
\begin{equation}
    G_t = \frac{\partial}{\partial{g}}( y_{t - 1} + \dot{y}_{t - 1}dt)
\end{equation}
\begin{equation}
    G_t = G_{t - 1} + \dot{G}_{t - 1}dt
\end{equation}
and 
\begin{equation}
    \dot{G}_{t - 1} = \dot{G}_{t - 2} + \ddot{G}_{t - 1} dt
\end{equation}

Using section 3.2, we get that
\begin{equation}
    \frac{\partial{f(x_{t-1}, g)}}{\partial{g}} = \frac{\psi_j w_j}{\sum_j \psi_j}x_{t-1}
\end{equation}

Hence: 
\begin{equation}
        \ddot{G}_{t - 1} =   \alpha(\beta(1 - G_{t - 2}) - \dot{G}_{t - 2}) +   \frac{\psi_j w_j}{\sum_j \psi_j}x_{t-1}
\end{equation}

and we get a similar relationship as Equation~\eqref{eq:final_diff}: 
\begin{equation}
    G_t = G_{t - 1} + (\dot{G}_{t - 2} + (\alpha(\beta(1 - G_{t - 2}) - \dot{G}_{t - 2}) +   \frac{\psi_j w_j}{\sum_j \psi_j}x_{t-1}) dt)dt
\end{equation}

Hence, $G_t$, similarly is dependent on $G_{t-1}, \dot{G}_{t - 2}, G_{t - 2}$. We can use a similar argument as with $w_i$ to show that $G_t$ is also numerically achievable. We have now shown that $y_t$, the output of the dynamical system defined by a DMP, is differentiable with respect to $w_i$ and $g$.

\subsection{Ablations}
\begin{figure}[th!]
\centering
\begin{subfigure}[b]{0.24\linewidth}
    \includegraphics[width=\linewidth]{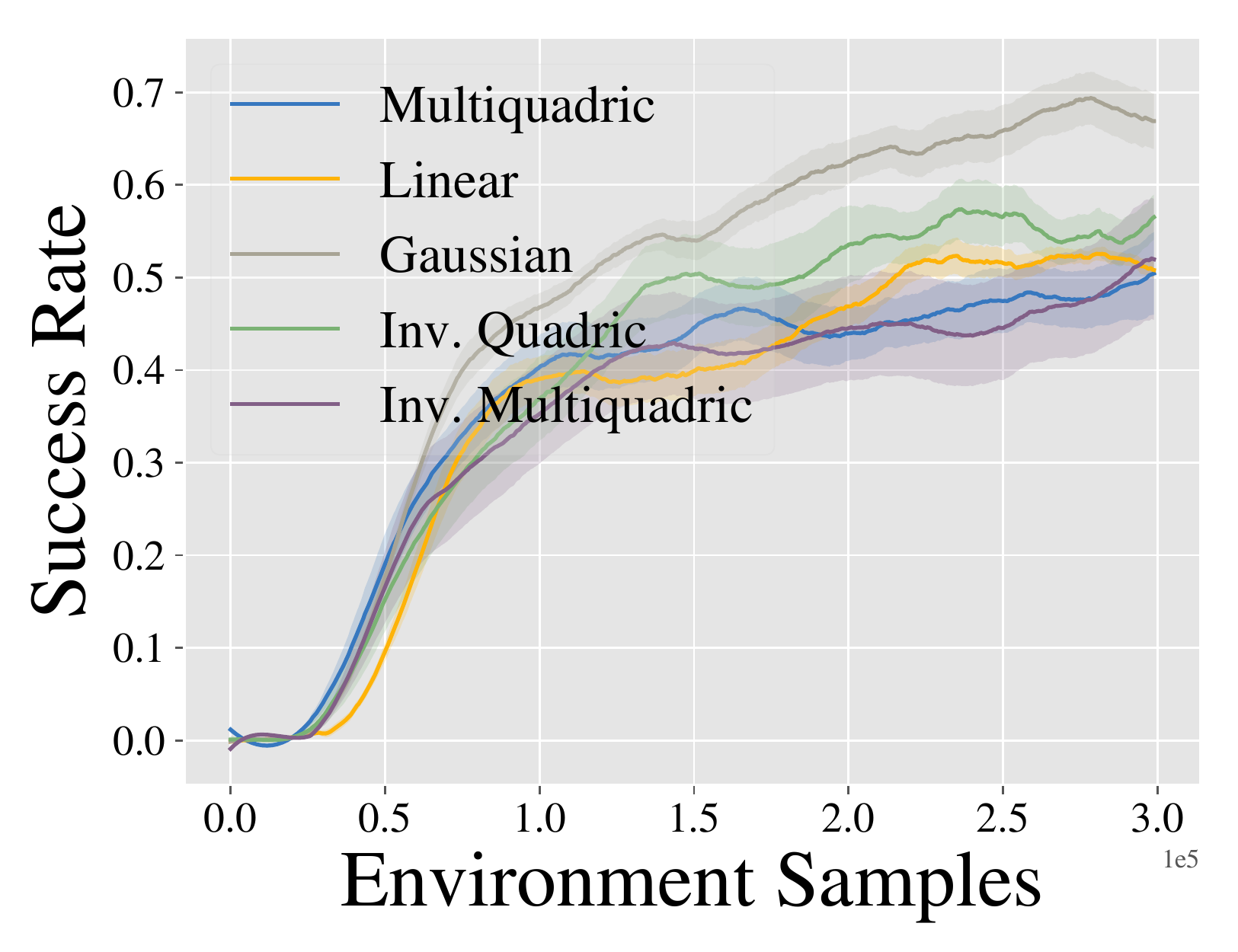}
    \vspace{-0.2in}
    \caption{\small RBF Kernels}
    \label{fig:t-rl-ablation-1}
\end{subfigure}
\begin{subfigure}[b]{0.24\linewidth}
    \includegraphics[width=\linewidth]{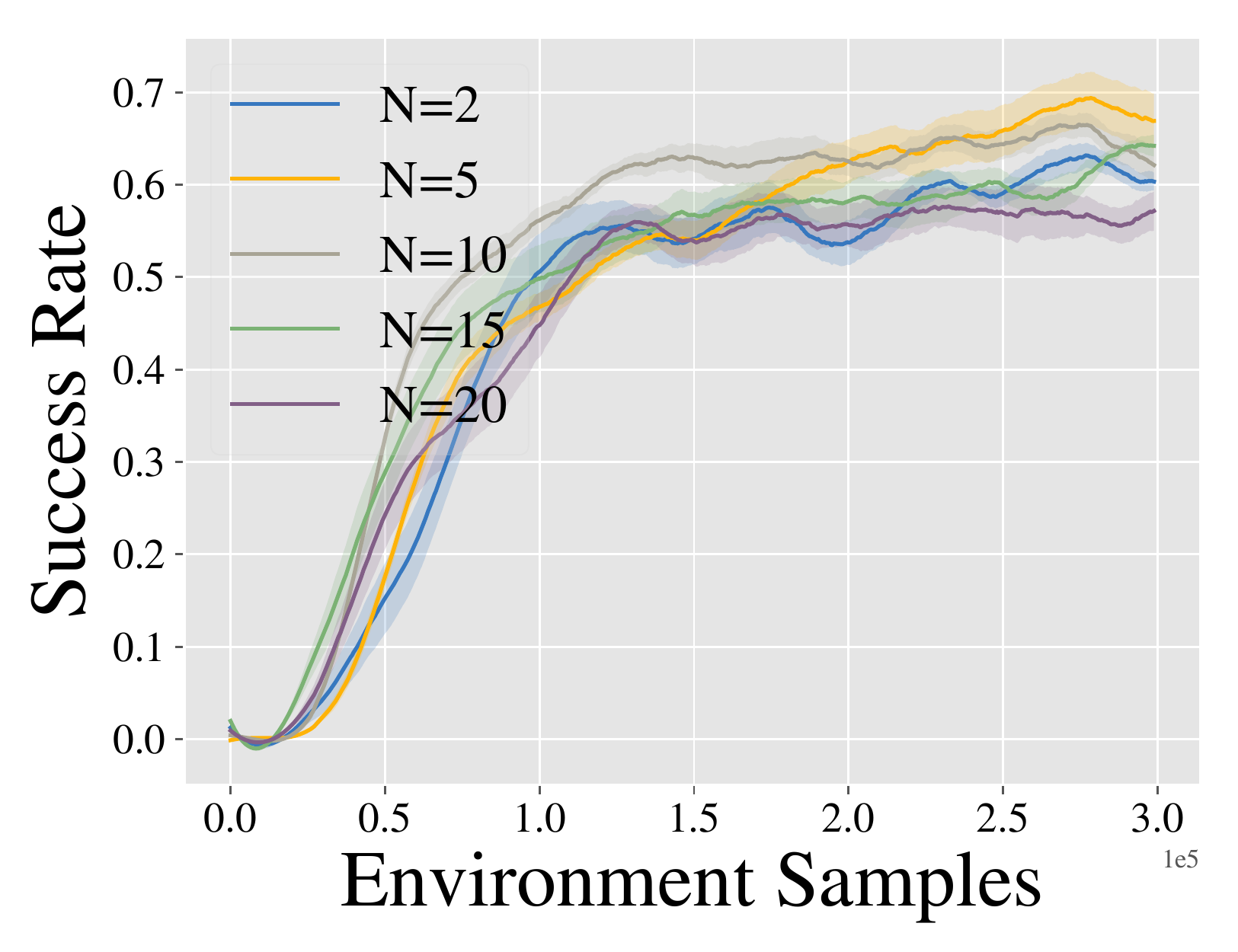}
    \vspace{-0.2in}
    \caption{\small \# of basis functions}
    \label{fig:t-rl-ablation-2}
\end{subfigure}
\begin{subfigure}[b]{0.24\linewidth}
    \includegraphics[width=\linewidth]{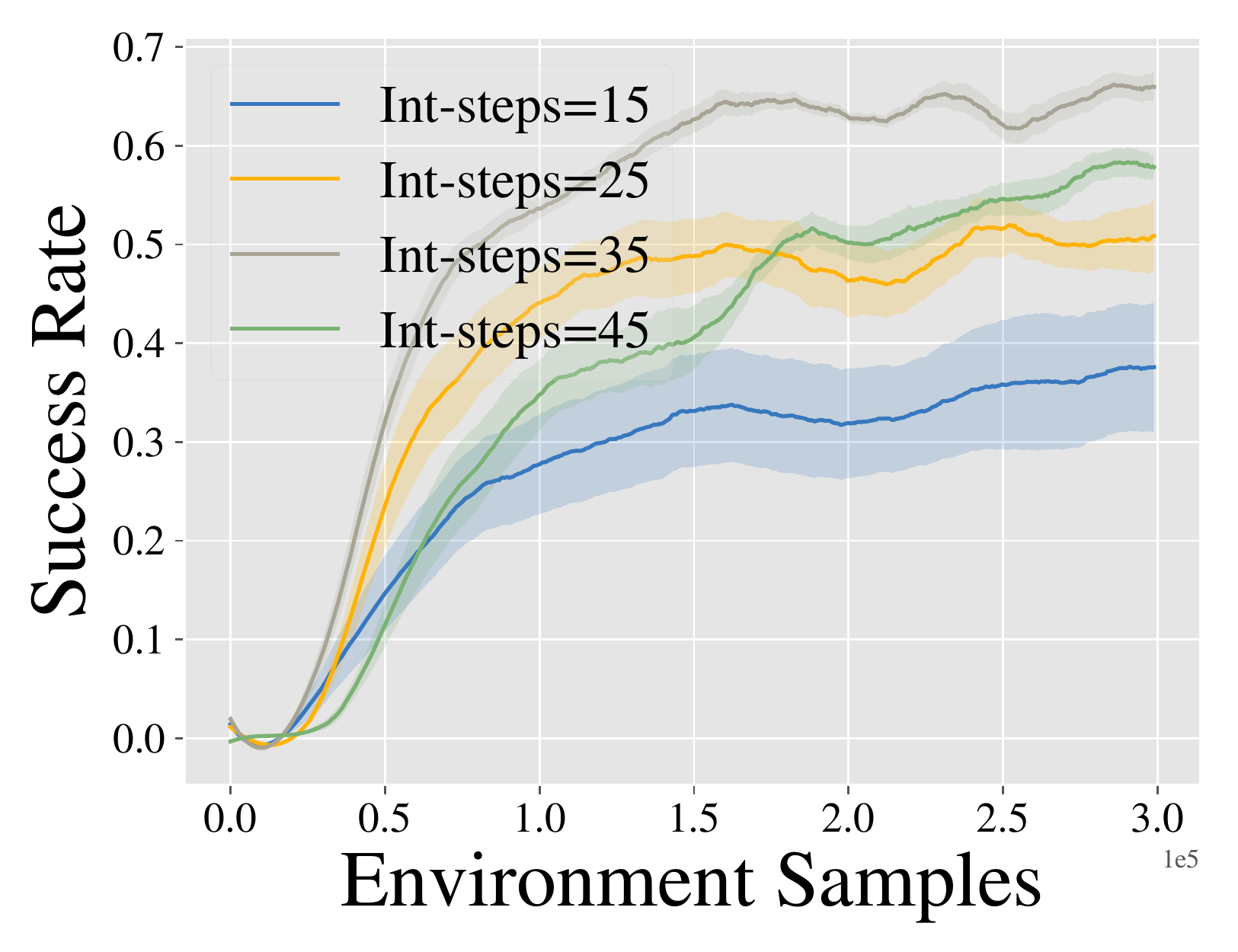}
    \vspace{-0.2in}
    \caption{\small Integration steps}
    \label{fig:t-rl-ablation-3}
\end{subfigure}
\begin{subfigure}[b]{0.24\linewidth}
    \includegraphics[width=\linewidth]{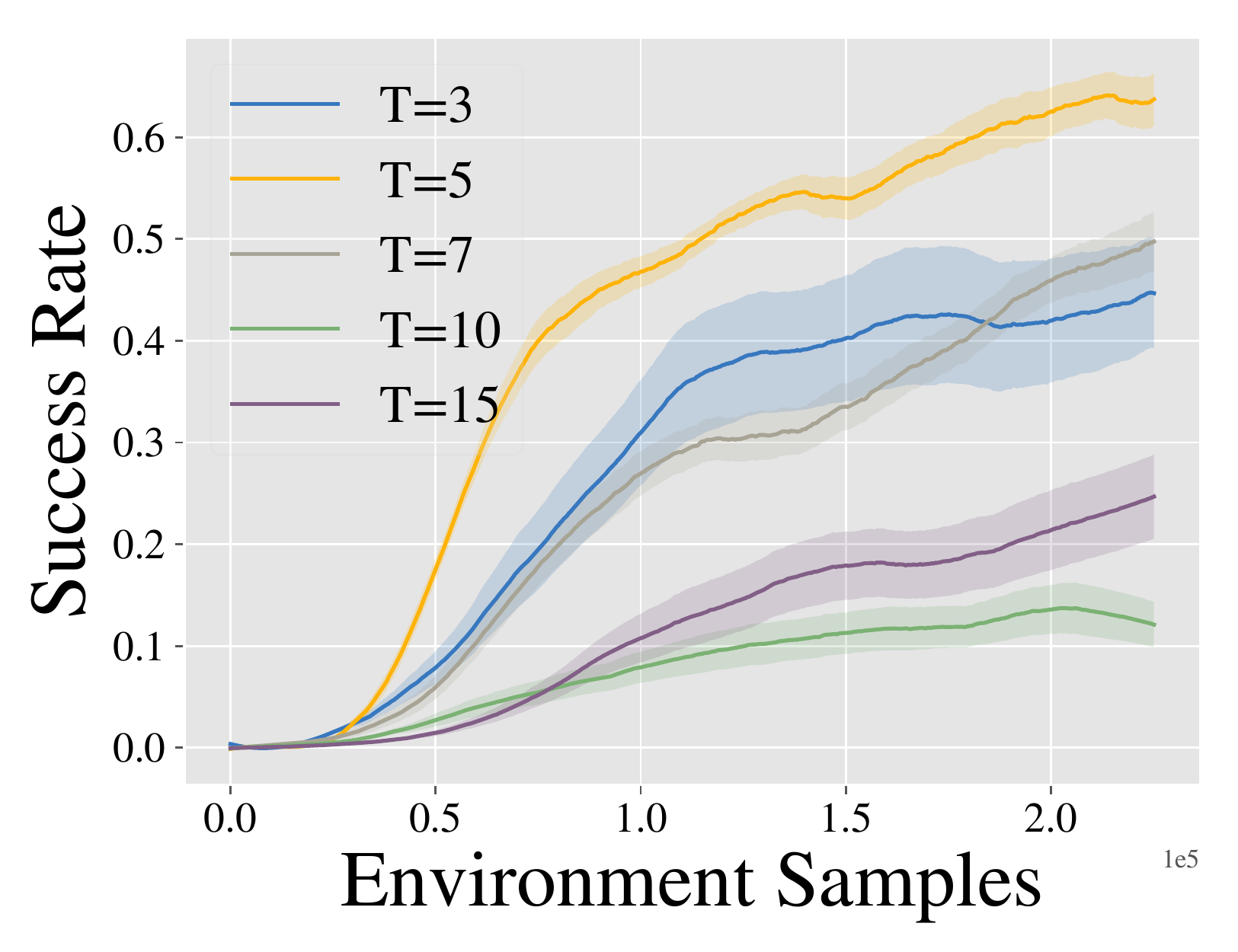}
    \vspace{-0.2in}
    \caption{\small Rollout length}
    \label{fig:t-rl-ablation-4}
\end{subfigure}
\vspace{-0.06in}
\caption{\small Ablation of \ours with respect to different hyperparameters in  the RL setup (throw). We ablate different choices of radial basis functions in (a). We ablate across number of basis functions, integration steps, and length of the NDP rollout in (b,c,d). Plots indicate that NDPs are fairly stable across a wide range of choices.}
\vspace{-0.06in}
\label{fig:abl-throw}
\end{figure}

\begin{figure}[ht!]
\centering
\begin{subfigure}[b]{0.32\linewidth}
    \includegraphics[width=\linewidth]{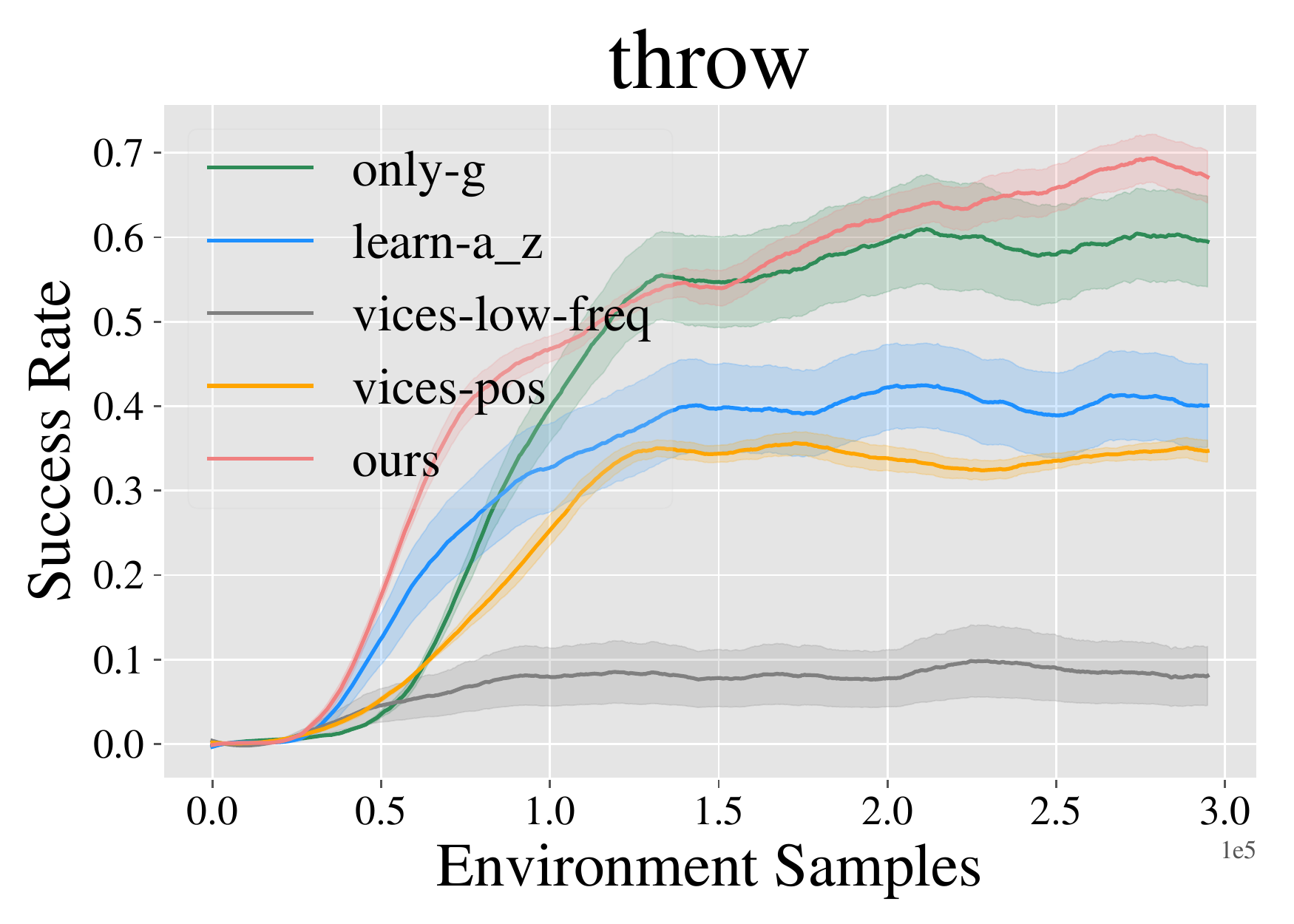}
    \vspace{-0.26in}
    \caption{\small Throwing}
    \label{fig:throw-rl-vices}
\end{subfigure}
\begin{subfigure}[b]{0.32\linewidth}
    \includegraphics[width=\linewidth]{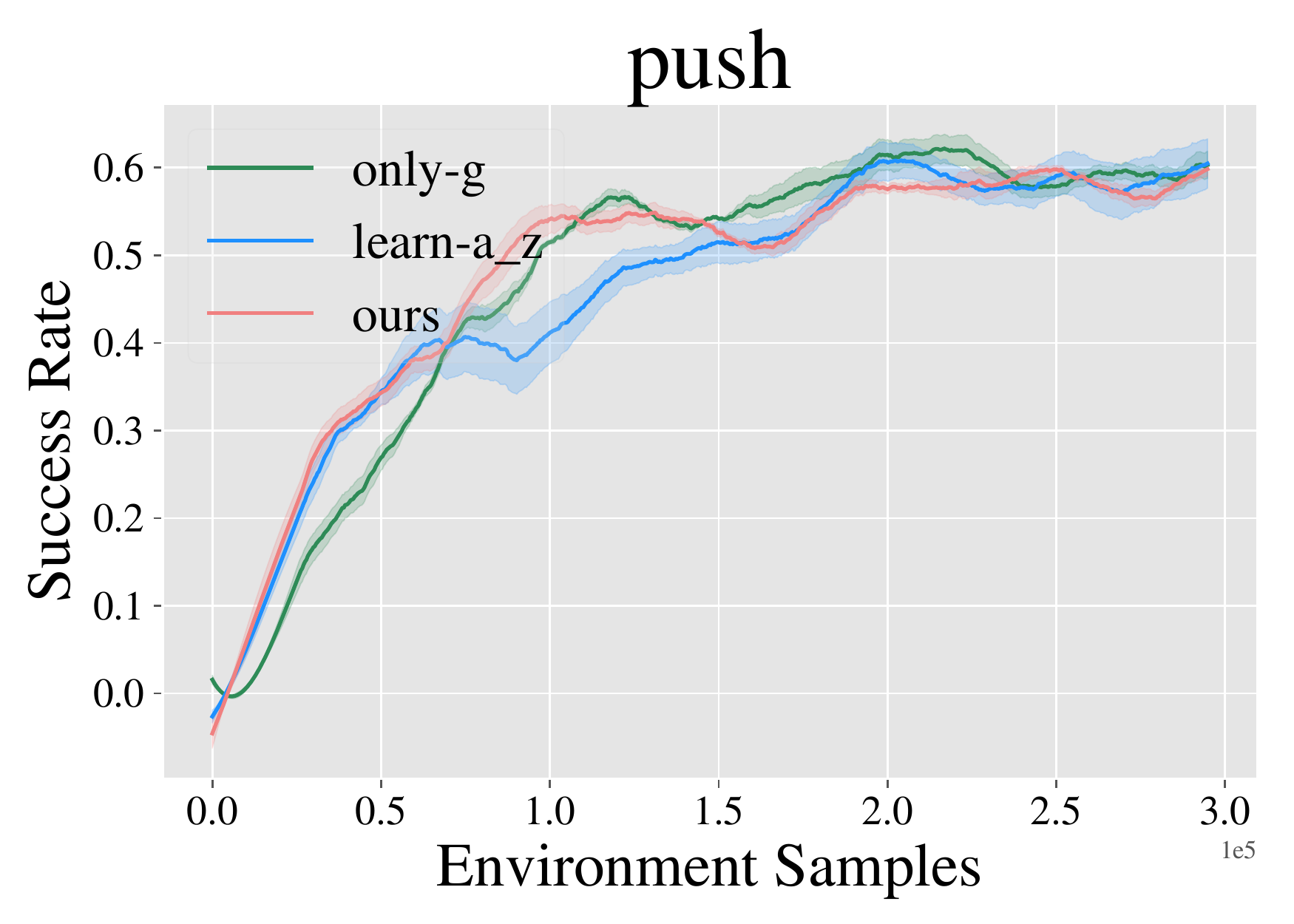}
    \vspace{-0.26in}
    \caption{\small Push}
    \label{fig:pick-rl-vices}
\end{subfigure}
\begin{subfigure}[b]{0.32\linewidth}
    \includegraphics[width=\linewidth]{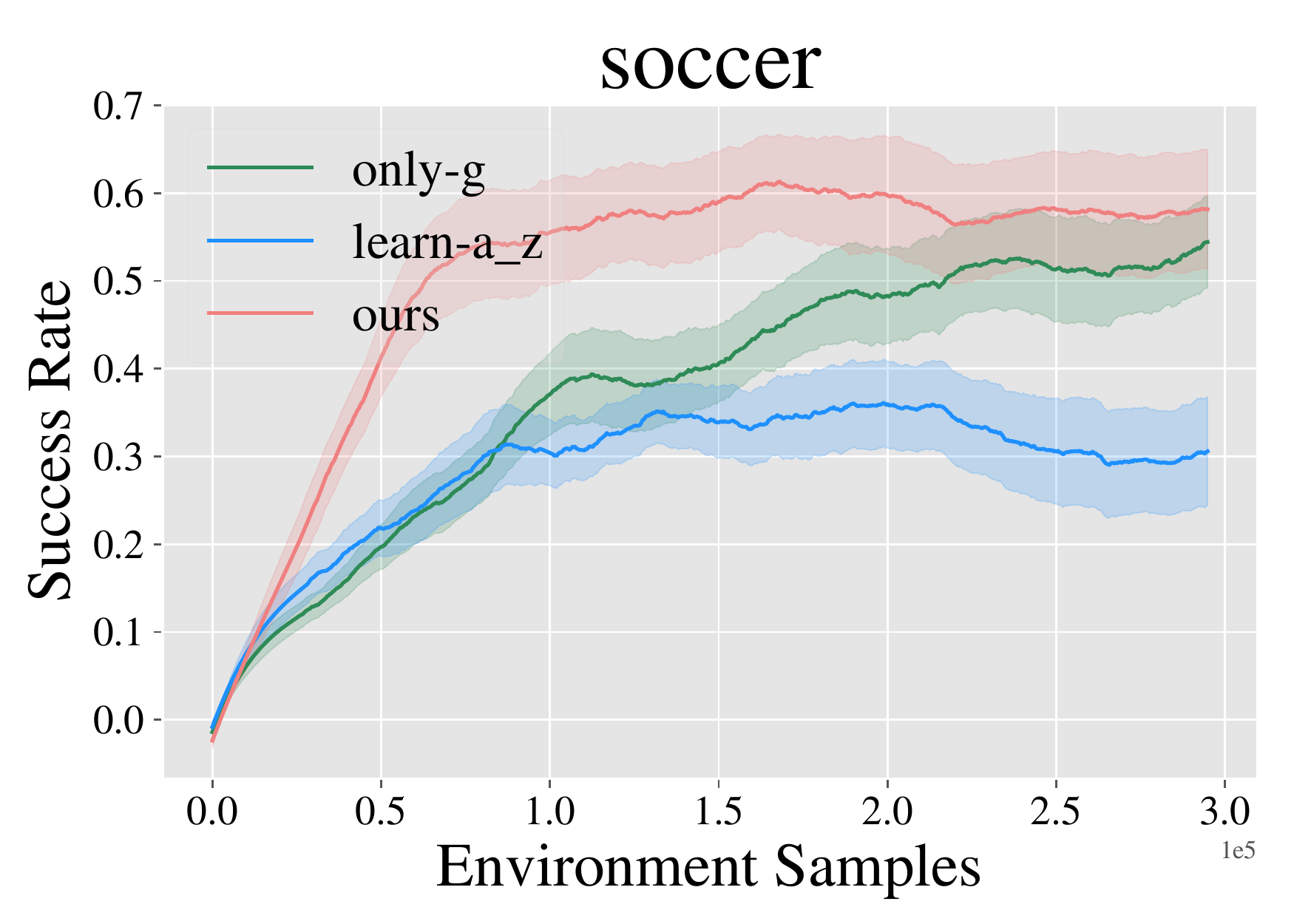}
    \vspace{-0.26in}
    \caption{\small Soccer}
    \label{fig:push-rl-vices}
\end{subfigure}
\begin{subfigure}[b]{0.32\linewidth}
    \includegraphics[width=\linewidth]{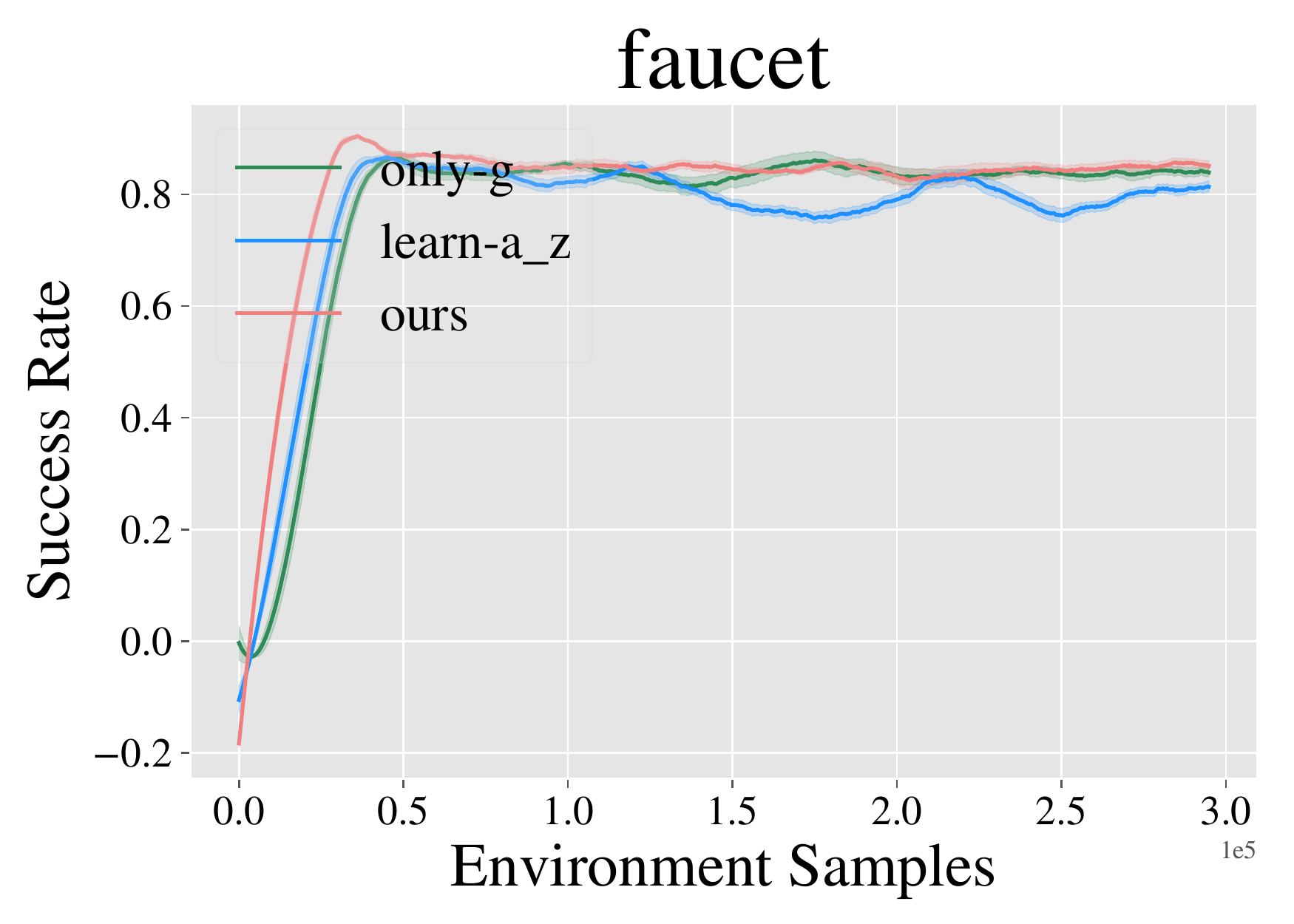}
    \vspace{-0.26in}
    \caption{\small Faucet Open}
    \label{fig:faucet-rl-vices}
\end{subfigure}
\begin{subfigure}[b]{0.32\linewidth}
    \includegraphics[width=\linewidth]{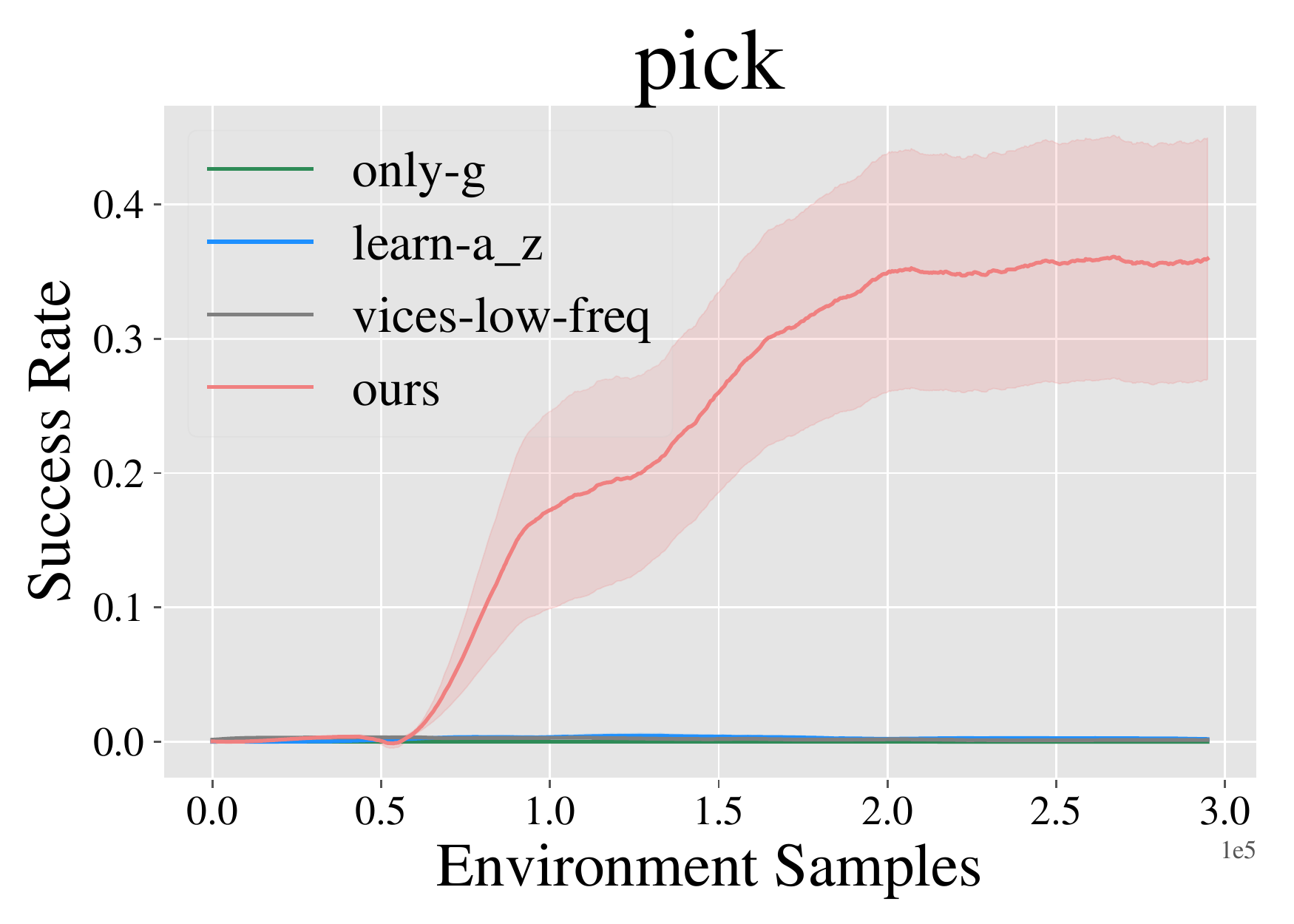}
    \vspace{-0.26in}
    \caption{\small Picking}
    \label{fig:soccer-rl-vices}
\end{subfigure}
\vspace{-0.01in}
\caption{\small Ablations for learning $\alpha$ (az) as well, only learning $g$ (only-g), VICES at a lower control frequency (vices-lower-freq)and VICES for Throw in end-effector space (vices-pos)}
\vspace{-0.06in}
\label{fig:abl-vices}
\end{figure}

We present ablations similar to the ones in our paper, using the Throwing task. The results are showin in Figure~\ref{fig:abl-throw}. We see that \ours show similar robustness across all variations. Secondly, we ran ablations with the forcing term set to $0$ and found it variant to be significantly less sample efficient than NDPs while converging to a slightly lower asymptotic performance. Finally, we ran ablation where $\alpha$ is also learned by the policy while setting $\beta = \frac{\alpha}{4}$ for critical damping. We see in Figure~\ref{fig:abl-vices} that \ours outperforms both settings where $\alpha$ is learnt and where the the forcing term $f$ is set to $0$. 

Additionally, we ran multiple ablations for the VICES baseline. We present a version of VICES  for throwing and picking tasks that acts in end-effector space instead of joint-space (we call this `vices-pos`)., as well 
e ran another version of VICES where the higher level policy runs at similar frequency as NDP which we call `vices-low-freq`. The results are presented in Figure~\ref{fig:throw-rl-vices} and Figure~\ref{fig:pick-rl-vices}. We found it to be less sample efficient and have a lower performance than \our. 

\subsection{Implementation Details}
\subsubsection*{Hyper-parameters and Design Choices}
We use default parameters and stick as closely as possible to the default code. In multi-action cases (for PPO-multi and NDP), we kept rollout length $k$ fixed for training and at inference time, one can sample arbitrarily high value for $k$ as demanded by the task setup. For reinforcement learning, we kept $k=5$ because the reward becomes too sparse if $k$ is very large. For imitation learning, this is not an issue, and hence, $k=300$ for learning from demonstration. 

In reinforcement learning setup for \our, we tried number of basis functions in [4, 5, 6, 7, 10] for each RL task. We fixed the number of integration steps per \our rollout to 35. We also tried $\alpha$ (as described in Section 2) values in [10, 15, 25]. NDP (ours), PPO~\cite{ppo}, PPO-Multi, VICES~\cite{vices2019martin} all use a similar 3-layer fully-connected network of hidden layer sizes of [100, 100] with tanh non-linearities. All these use PPO~\cite{ppo} as the underlying RL optimizer. For Dyn-E~\cite{whitney2019dynamics}, we used off-the-shelf architecture because it is based on off-policy RL and doesn't use PPO.

Hyper-parameters for underlying PPO optimization use off-the-shelf without any further tuning from PPO~\cite{ppo} implementation in~\citet{pytorchrl} as follows:

\begin{table}[ht]
\centering
\begin{tabular}{lc}
\toprule
\textbf{Hyperparameter} & \textbf{Value}\\
\midrule
Learning Rate &$3 \times 10^{-4}$ \\
Discount Factor & 0.99 \\ 
Use GAE & True \\
GAE Discount Factor & 0.95 \\ 
Entropy Coefficient & 0 \\ 
Normalized Observations & True \\ 
Normalized Returns & True \\ 
Value Loss Coefficient & 0.5 \\ 
Maximum Gradient Norm & 0.5 \\ 
PPO Mini-Batches & 32 \\ 
PPO Epochs & 10 \\ 
Clip Parameter & 0.1 \\ 
Optimizer & Adam \\
Batch Size & 2048 \\
RMSprop optimizer epsilon &  $10^{-5}$ \\ 
\bottomrule
\end{tabular}
\vspace{0.1cm}
\end{table}

\subsubsection*{Environments}
For Picking and Throwing, we adapted tasks from \citet{ghosh2017divide} (\url{https://github.com/dibyaghosh/dnc}. We modified these tasks to enable joint-angle position control. For other RL tasks, we used the Meta-World environment package \cite{yu2019meta} (\url{https://github.com/rlworkgroup/metaworld}). Since VICES \cite{vices2019martin} operates using torque control, we modified Meta-World environments to support torque-based Impedance control. 

Further, as mentioned in the Section 3.4 and 4, \our and PPO-multi are able to operate the robot at a higher frequency than the world. Precisely, frequency is $k$-times higher where $k=5$ is the \our rollout length (described in Section~3.2). Even though the robot moves at higher frequency, the environment/world state is only observed at normal rate, i.e., once every $k$ robot steps and the reward computation at the intermediate $k$ steps only use stale environment/world state from the first one of the $k$-steps. For instance, if the robot is pushing a puck, the reward is function of robot as well as puck's location. The robot will knows its own position at every policy step but will have access to stale value of puck's location only from actual environment step (sampled at a lower frequency than policy steps, specifically 5x less).  We implemented this for all 50 Meta-World environments as well as Throwing and Picking.

\subsubsection*{Codebases: NDPs (ours) and Baselines}
Our code can be found at: \url{https://shikharbahl.github.io/neural-dynamic-policies/}. Our algorithm is based on top of Proximal Policy Optimization (PPO) \cite{ppo} from \url{https://github.com/ikostrikov/pytorch-a2c-ppo-acktr-gail} \cite{pytorchrl}.
Additionally, we use code from \citet{whitney2019dynamics} (DYN-E): \url{https://github.com/willwhitney/dynamics-aware-embeddings}. For our implementation of VICES \cite{vices2019martin}, we use the controllers provided them in \url{https://github.com/pairlab/robosuite/tree/vices_iros19/robosuite} and overlay those on our environments.

\end{document}